\pdfoutput=1

\documentclass[11pt]{article}

\usepackage[final]{acl}

\usepackage{times}
\usepackage{latexsym}

\usepackage[T1]{fontenc}

\usepackage[utf8]{inputenc}

\usepackage{microtype}

\usepackage{inconsolata}

\usepackage{graphicx}
\usepackage{hyperref}

\usepackage{booktabs}
\usepackage{array}
\usepackage{makecell}
\usepackage{xcolor}
\usepackage{enumitem}
\usepackage{amssymb}
\usepackage{amsmath}
\usepackage{forest}
\usepackage{todonotes}

\usepackage{tablefootnote}
\newcommand{\ttt}[1]{\texttt{#1}}
\definecolor{darkgreen}{RGB}{0,100,0}

\newcommand{\comprehension}{\texttt{X$\rightarrow$model}}
\newcommand{\generation}{\texttt{model$\rightarrow$X}}
\newcommand{\WordTranslationAbbreviation}{WT}
\newcommand{\WordTranslationInContextAbbreviation}{WTWC}
\newcommand{\FillInTheBlankAbbreviation}{TCLM}
\newcommand{\BagOfWordsMachineTranslationAbbreviation}{BOW MT}
\newcommand{\ProjectName}{ChiKhaPo}
\newcommand{\ayaOne}{\texttt{aya-101}}
\newcommand{\ayaTwo}{\texttt{aya-23-8b}}
\newcommand{\bloom}{\texttt{bloomz-7b1-mt}}
\newcommand{\falcon}{\texttt{falcon-7b-instruct}}
\newcommand{\gemma}{\texttt{gemma-2b-it}}
\newcommand{\llama}{\texttt{Llama-3.1-8B-Instruct}}

\title{
\protect\raisebox{-0.1\height}{%
\protect\includegraphics[height=1cm]{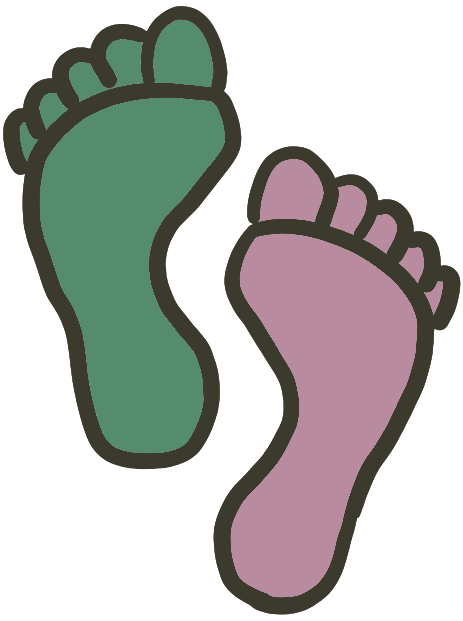}}
\hspace{-0.3em}
\ProjectName : A Large-Scale Multilingual Benchmark for Evaluating Lexical Comprehension and Generation in Large Language Models
}

\author{
    Emily Chang\textsuperscript{1} and Niyati Bafna\textsuperscript{2} \\
    \textsuperscript{1} Toyota Technological Institute at Chicago;\\\textsuperscript{2} Johns Hopkins University, Center for Language and Speech Processing
}

\begin{document}
\maketitle
\begin{abstract}
Existing benchmarks for large language models (LLMs) are largely restricted to high- or mid-resource languages, and often evaluate performance on higher-order tasks in reasoning and generation.
However, plenty of evidence points to the fact that LLMs lack basic linguistic competence in the vast majority of the world’s $3800+$ written languages.
We introduce \ProjectName, consisting of eight subtasks of varying difficulty designed to evaluate the lexical comprehension and generation abilities of generative models. 
\ProjectName~ draws on existing lexicons, monolingual data, and bitext, and provides coverage for $2700+$ languages for two word-translation-based subtasks, surpassing any existing benchmark in terms of language coverage. We further show that six SOTA models struggle on our benchmark, and discuss the factors contributing to performance scores, including language family, language resourcedness, task, and comprehension versus generation directions. 
With \ProjectName, we hope to enable and encourage the massively multilingual benchmarking of LLMs.\footnote{We release our dataset, code for our experiments, and package for running our benchmark. \\ Dataset: \href{https://huggingface.co/datasets/ec5ug/chikhapo}{huggingface.co/datasets/ec5ug/chikhapo}\\
Code: \href{https://github.com/ec5ug/chikhapo/}{github.com/ec5ug/chikhapo/}\\
Python package: \href{https://pypi.org/project/chikhapo/}{pypi.org/project/chikhapo/}}

\end{abstract}

\section{Introduction}

Benchmarks are crucial for not only measuring but steering progress in NLP \citep{ruder2021challenges}. 
While LLMs are capable of impressive feats of complex reasoning and content generation \citep{deepseekai2025deepseekr1incentivizingreasoningcapability, bercovich2025llama, chen2025researchlearningreasonsearch}, these capabilities are restricted to a few dozen high-resource languages (HRLs) among 3800+ written languages and dialects in the world \citep{aji-etal-2022-one, ebrahimi-etal-2022-americasnli}. %
The availability of evaluation benchmarks reflects this problem, with the most multilingual of these at the time of writing being FLORES+ \citep{nllb-24}, which tests machine translation (MT) for $212$ languages.
For the rest of the world's languages, we have no way to assess even basic LLM capabilities. %
\begin{figure}
    \centering
    \includegraphics[width=\linewidth]{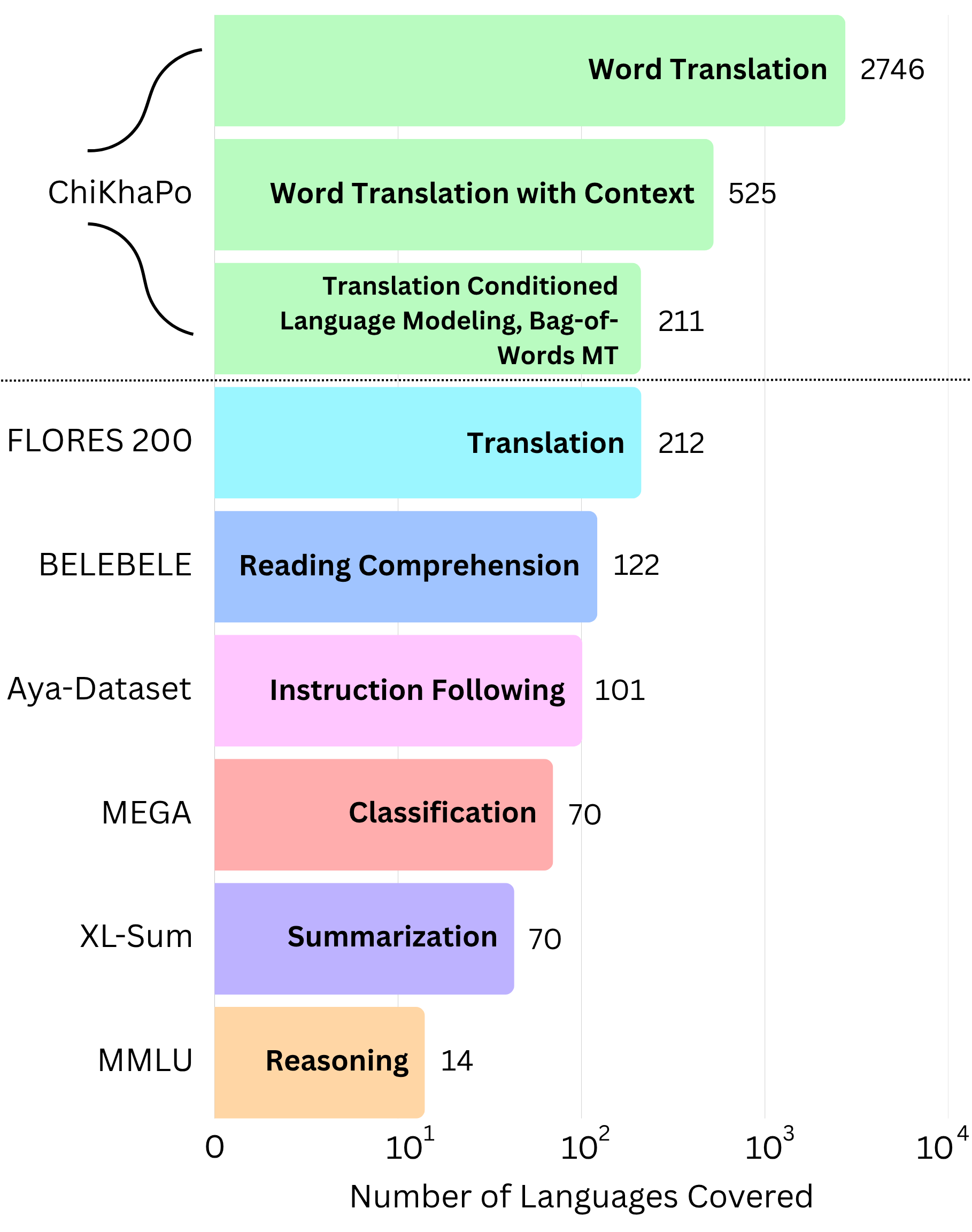}
    \caption{\ProjectName~ evaluates basic lexical competence with several tasks, covering an order of magnitude more languages than existing multilingual benchmarks.} 
    \label{fig:current_prior_benchmark_language_coverage}
\end{figure}

We introduce \ProjectName, a benchmark that measures basic lexical comprehension and generation abilities in LLMs on a massively multilingual scale.\footnote{The name is inspired by the Hokkien saying that progress is made step-by-step: \emph{chit kha-po, chit kha-in}.} %
\ProjectName~ includes $4$ tasks $\times$ $2$ evaluation directions. The tasks provide various perspectives on lexical competence, and the evaluation directions measure model ability for lexical \textit{comprehension} (\comprehension) and \textit{generation} (\generation) per task. 
The tasks include 1) \textbf{word translation (\WordTranslationAbbreviation)}, involving direct prompting for word translation, 2) \textbf{word translation with context (\WordTranslationInContextAbbreviation)}, involving direct prompting for word translation with source context cues, 3) \textbf{translation-conditioned language modeling (\FillInTheBlankAbbreviation)}, involving next word generation given source and target language context in a natural machine translation setting, and 4) \textbf{bag-of-words machine translation (\BagOfWordsMachineTranslationAbbreviation)}, involving word generation as part of a sentence-level translation task.
Each task and direction is evaluated at the word level for a target language.\footnote{In this paper, the term ``target language'' refers to the language being evaluated, which may not be the language being generated. We use the terms ``source-side'' and ``target-side'' instead to refer to the input and output languages of the model.}

\ProjectName's subtasks make use of existing lexicons, monolingual data, and bitext. In particular, \WordTranslationAbbreviation~ relies solely on lexicons, and \WordTranslationInContextAbbreviation~ additionally requires monolingual data.
Both resources are widely available for many languages \citep{kamholz-etal-2014-panlex,imanigooghari-etal-2023-glot500};  
thus, \ProjectName~ covers $2700+$ and $500+$ languages for these tasks respectively, which surpasses the coverage of any existing benchmark (see \autoref{fig:current_prior_benchmark_language_coverage}).
We also show that performance on \WordTranslationAbbreviation~ is correlated with sentence-level MT performance, 
providing a simple proxy in the absence of bitext.

We evaluate 6 state-of-the-art multilingual LLMs on our benchmark. We provide an analysis of the factors affecting their performance, such as subtask, language resourcedness, and language family, and thus highlight several avenues of focus for improving the broad multilingual competence of LLMs.

\ProjectName~ aims to fill two important gaps in current benchmarks. First, it evaluates \emph{core lexical abilities} in LLMs and allows us to track the ``atomic'' word-level competence of an LLM in a given language.
Second, it does so on a \emph{massively multilingual scale}. %
With this work, we hope to draw attention to the pressing issue of language inequity in NLP \citep{joshistate}, and promote the massively multilingual evaluation of LLMs. %

\section{Related Work}

\paragraph{LLM evaluation benchmarks}
Most existing benchmarks that LLMs are evaluated on focus on English and other high-resource languages \citep{grattafiori2024llama, aryabumi2024aya23openweight, qwen2025qwen25technicalreport}.
Popular benchmark suites include BIG-Bench \citep{srivastava2023beyond}\textemdash a collection of $200$ tasks testing various kinds of comprehension and generation\textemdash and HELM \citep{liang2023holistic}, a framework that standardizes LLM reasoning and generation and provides metrics beyond accuracy (e.g. calibration). 
Datasets such as XNLI \citep{conneau2018xnlievaluatingcrosslingualsentence} and XCOPA \citep{ponti2020xcopa} measure reasoning skills with classification-style tasks, whereas natural language generation is evaluated with datasets for summarization, machine translation, and instruction following, such as XL-SUM \citep{hasan-etal-2021-xl}, FLORES+ \citep{nllb-24}, BOUQuET \citep{andrews-etal-2025-bouquet,omnilingual2026}, and the Aya Evaluation Suite \citep{singh2024aya}. 

In Appendix \autoref{tab:prior_multiling_benchmark_coverage}, we list 20+ commonly used datasets in LLM multilingual benchmarking.
These datasets test a collection of relatively complex tasks and cover a limited number of languages.

\paragraph{Lexical evaluation}
\citet{mccarthy-2002-lexical} first introduced \emph{lexical substitution}, the task of choosing an appropriate substitute for a word given a context to test word sense disambiguation systems. Prior lexical substitution benchmarks are overwhelmingly English \citep{mccarthy-navigli-2007-semeval, kremer-etal-2014-substitutes, lee-etal-2021-swords} These benchmarks are small and manually designed. %

In implementing \ProjectName, we adopted the approach of \citet{mihalcea-etal-2010-semeval} who coined the term \textit{cross-lingual lexical substitution}, and evaluated lexical understanding using translations rather than paraphrases.
\citet{martinez_et_al_vocab_tests_2024} uses expert-designed vocabulary tests to perform a fine-grained evaluation of LLMs; however, the benchmark is limited to English and Spanish.

As far as we know, our work is the first to design a lexical competence benchmark with a massively multilingual scope using existing resources. %

\section{Tasks} \label{sec:dataset_description}

\ProjectName's suite of tasks centers on lexical semantics, the branch of semantics concerned with word meaning. A word has two meanings: grammatical and lexical. While grammatical meaning refers to the word's function in a language (e.g. plurality, tense), we focus on the word's \textit{lexical meaning}, or the denotative meaning of the base word \citep{Pustejovsky_2016}. %

\newcolumntype{P}[1]{>{\raggedright\arraybackslash}p{#1}}
\begin{table*}[hbtp]
    \centering
    \small
    \begin{tabular}{P{3cm} P{5.5cm} P{5.5cm}}
        \toprule
        \multicolumn{1}{c}{\textbf{Task}} & 
        \multicolumn{1}{c}{\textbf{Comprehension: \texttt{X $\rightarrow$ model}}} & 
        \multicolumn{1}{c}{\textbf{Generation: \texttt{model $\rightarrow$ X}}} \\
        \midrule
        Word Translation & 
        \begin{minipage}[t]{\linewidth}
            \textbf{Input}: Translate the following text from Malay to English: \texttt{ujan}.

            \noindent\dotfill

            \textbf{Correct Output}: \ttt{rain}

            \noindent\dotfill
            
            \textbf{Model Output}: \texttt{rain}

            \noindent\dotfill
            
            \textbf{Score}: \\
            \ttt{scores[``ujan''] += 1}
        \end{minipage} & 
        \begin{minipage}[t]{\linewidth}
            \textbf{Input}: Translate the following text from English to Afrikaans: \texttt{attacked.}

            \noindent\dotfill
            
            \textbf{Correct Output}: \ttt{aangeval}

            \noindent\dotfill
            
            \textbf{Model Output}: \ttt{aangeval}

            \noindent\dotfill
            
            \textbf{Score}: \\
            \ttt{scores[``aangeval''] += 1}
        \end{minipage} \\
        \midrule
        Word Translation with Context &
        \begin{minipage}[t]{\linewidth}
            \textbf{Input}: In `Minonke konam phoro isi sonturi aghaipo aro anang pen Jisu yok honsi kido, aro alok hel, labadi chiklik hel aro ajat jat kachiplang theksi, anali chiphere detno, aro pulo, ``Khanangsi labang arlengpo Arnam Aso kido . '' ', the word `\ttt{kido}' means \_\_\_\_ in English.
            
            \noindent\dotfill

            \textbf{Correct Output}: \ttt{letter}

            \noindent\dotfill
            
            \textbf{Model Output}: \ttt{child}

            \noindent\dotfill
            
            \textbf{Score}:\\
            \ttt{scores[``kido''] += 0}
        \end{minipage} &
        \begin{minipage}[t]{\linewidth}
            \textbf{Input}: In `After the match, King of Clay said, ``I am just excited about being back in the final rounds of the most important events. I am here to try to win this.'' ', the word `win' means \_\_\_\_ in Basque. \\ \\

            \noindent\dotfill

            \textbf{Correct Output}: \ttt{aurrea hartu}
            
            \noindent\dotfill
            
            \textbf{Model Output}: \ttt{ganar}

            \noindent\dotfill
            
            \textbf{Score}: \\
            \ttt{scores[``aurrea hartu''] += 0}
        \end{minipage} \\
        \midrule
        Translation-Conditioned Language Modeling &
        \begin{minipage}[t]{\linewidth}
            \textbf{Input}: Translate the following text into English: \\
            Dyula: \texttt{Aka dugutaga se'n fei, Iwasaki ye \textcolor{blue}{kassara} chaman le sôrô.} \\ \\
            English: \texttt{During his trip, Iwasaki }

            \noindent\dotfill

            \textbf{Reference Translation}: During his trip, Iwasaki \ttt{\textcolor{blue}{ran}} into trouble on many occasions. \\
            
            \noindent\dotfill

            \textbf{Model Output}:\\
            $\mathrm{P} [\mathtt{ran} \mid \mathtt{input}] = 0.567$

            \noindent\dotfill

            \textbf{Score}: \\
            \ttt{scores[``kassara''] += 0.567}
        \end{minipage} &
        \begin{minipage}[t]{\linewidth}
            \textbf{Input}: Translate the following text into Iloko. \\English: \texttt{"We now have 4-month-old mice that are non-diabetic that used to be diabetic," he added.} \\Iloko: \texttt{"Addaan kami ti 4-a-bulan a}

            \noindent\dotfill

            \textbf{Reference Translation}: ``Addaan kami ti 4-a-bulan a \ttt{\textbf{babbao}} a dati ket diabetic ngem saan itan,'' nainayonna.
            
            \noindent\dotfill

            \textbf{Model Output}:\\
            $\mathrm{P}[\mathtt{babbao} \mid \mathtt{input}] = 0.351$

            \noindent\dotfill

            \textbf{Score}: \\
            \ttt{scores[``babbao''] += 0.351}
        \end{minipage} \\
        \midrule
        Bag-of-Words Machine Translation &
        \begin{minipage}[t]{\linewidth}
            \textbf{Input}: Translate into English: \texttt{\textcolor{red}{Los trabalhadors} devon \textcolor{blue}{sovent} \textcolor{violet}{obtenir} \textcolor{teal}{l’aprobacion} de \textcolor{brown}{sos superiors}}.

            \noindent\dotfill

            \textbf{Reference Translation}: \texttt{\textcolor{red}{Workers} must \textcolor{blue}{often} \textcolor{violet}{get} \textcolor{brown}{their superiors'} \textcolor{teal}{approval}} \\

            \noindent\dotfill

            \textbf{Model Output}: \texttt{Workers often need to obtain their superiors' approval}

            \noindent\dotfill
            
            \textbf{Score}: \\
            \ttt{scores[\textcolor{red}{``los''}] += 1} \\
            \ttt{scores[\textcolor{red}{``trabalhadors''}] += 1} \\
            \ttt{scores[``devon"] += 0} \\
            \ttt{scores[\textcolor{blue}{``sovent''}] += 1} \\
            \ttt{scores[\textcolor{violet}{``obtenir''}] += 1} \\
            \ttt{scores[\textcolor{teal}{``l’aprobacion''}] += 1} \\
            \ttt{scores[``de''] += 0} \\
            \ttt{scores[\textcolor{brown}{``sos''}] += 1} \\
            \ttt{scores[\textcolor{brown}{``superiors''}] += 1}
        \end{minipage} &
        \begin{minipage}[t]{\linewidth}
            \textbf{Input}: \texttt{Workplace harmoney is crucial} \\ \\

            \noindent\dotfill

            \textbf{Reference Translation}: Ukusebenza ngokubambisana endaweni yokusebenzela kubalulekile
            
            \noindent\dotfill
            
            \textbf{Model Output}: \texttt{Ukuzwana endaweni yokusebenza kubalulekile}

            \noindent\dotfill
            
            \ttt{scores[``ukusebenza''] += 0} \\
            \ttt{scores[``ngokubambisana''] += 0} \\
            \ttt{scores[``endaweni''] += 1} \\
            \ttt{scores[``yokusebenzela''] += 1} \\
            \ttt{scores[``kubalulekile''] += 1} \\
        \end{minipage} \\
        \bottomrule
    \end{tabular}
    \caption{Example task prompts, model outputs, and vocabulary-based scores. These scores are aggregated over target language words as per \autoref{sec:dataset_description}. For the \texttt{X $\rightarrow$ model} direction, English output words are aligned to input language words for scoring; alignment is shown via coloring.}
    \label{tab:example_task_prompts}
\end{table*}

\begin{table*}[hbtp]
    \centering
    \small
    \resizebox{\linewidth}{!}{
    \begin{tabular}{c|c c|c c|c c}
        \toprule
        & \multicolumn{2}{c|}{\textbf{Vocabulary Size}} & \multicolumn{2}{c|}{\textbf{Total Word Count}} & \multicolumn{2}{c}{\makecell{\textbf{Number}\\ \textbf{of Languages}}} \\ %
        \midrule
        \textbf{Task} & \ttt{\comprehension} & \texttt{\generation} & \texttt{\comprehension} & \texttt{\generation} & \texttt{\comprehension} & \texttt{\generation} \\ %
        \midrule
        \WordTranslationAbbreviation & 4.8K $\pm$ 39K & 4.8K $\pm$ 39K & 4.8K $\pm$ 39K & 9.4K $\pm$ 74K & 2746 & 2746 \\ \midrule
        \WordTranslationInContextAbbreviation & 2.4K $\pm$ 4.8K & 8.2K $\pm$ 18K & 410K $\pm$ 630K & 9700K $\pm$ 19000K & 525 & 525 \\ \midrule
        \FillInTheBlankAbbreviation & 7.4K $\pm$ 11K & 6.8K $\pm$ 1.6K & 90K $\pm$ 140K & 21K $\pm$ 5.4K & 211 & 211 \\ \midrule
        \BagOfWordsMachineTranslationAbbreviation  & 7.4K $\pm$ 11K & 6.8K $\pm$ 1.6K & 90K $\pm$ 140K & 21K $\pm$ 5.4K & 211 & 211 \\
        \bottomrule
    \end{tabular}}
    \caption{Vocabulary size: Mean and standard deviation of the number of unique words over languages per subtask. Total word count: Mean and standard deviation of total word count per language, relevant for tasks where a single word can be tested in multiple contexts. Vocabulary size and total word count are expressed in the thousands (K). Large standard deviations are caused by HRL outliers.}
    \label{tab:dataset_breakdown_by_lang}
\end{table*}

Given the English-centricness of LLMs \citep{wendler2024llamasworkenglishlatent}, we treat the model's ability to translate a word \textit{into English} as a proxy for its comprehension of the word (\comprehension), and its ability to generate the word when translating \textit{from English} as a proxy for its generation capability for that word (\generation). 

We design $8$ subtasks: $4$ tasks in two directions each (\comprehension, \generation), aimed at examining various facets of lexical capabilities in LLMs, and described in detail below. 
For all subtasks, we calculate our metrics over target language words (i.e. not English words $w_{(i)}^E$). More specifically, we assign the $i^{th}$ word $w_{(i)}^X$ in the target language $X$ a score $s(w_{(i)}^X) \in [0, 1]$.
We calculate aggregate scores for language $L_{(\lambda)}$ over its vocabulary and for a model $M_{(\kappa)}$ over languages:
\begin{align*}
    s(L_{(\lambda)}) &= \bigg( \frac{1}{|V|} \sum_{i=1}^{|V|} s(w_{(i)}^X) \bigg) \times 100\% \\
    s(M_{(\kappa)}) &= \frac{1}{|L|} \sum_{\lambda=1}^{|L|} s(L_{(\lambda)})
\end{align*}

We describe below how word scores $s(w_{(i)}^X)$ are calculated in each of our $8$ subtasks.
\autoref{tab:example_task_prompts} displays example inputs, outputs, and associated scores for each subtask. We also provide more examples per task in \autoref{sec:classification_heuristics}. We list dataset sizes and number of supported languages for each subtask in \autoref{tab:dataset_breakdown_by_lang}. %

\subsection{Word Translation}

In this task, we directly prompt a model to translate an input word either into or out of English for every term within a bilingual lexicon. %

\subsubsection{Scoring} 
For a given model output, we check for equivalence against all translation equivalents of the source word from our lexicon $\Xi$, using $\Xi(w_{(i)})$ to refer to the set of equivalents of $w_{(i)}$. %
Note that requiring answers to be an exact match to lexicon translations is unfairly strict, as the model may output a different morphological form of the correct equivalent or extraneous text around the correct answer. Given these considerations, we use additional string-matching heuristics, such as \ttt{inflection} and \ttt{substring}, among others, to determine if the model output is equivalent to the reference. We also check for \ttt{synonym}y using the English WordNet \citep{miller-1994-wordnet} in the \comprehension~ direction. See \autoref{sec:classification_heuristics} for more examples and an analysis of the false positive and negative rates of these heuristics across tasks. %

\paragraph{\comprehension} Given a word to translate $w_{(i)}^X$ and the model prediction $\hat w_{(i)}^E$, we compute the binary correctness variable  $\alpha_{\mathrm{\comprehension}}^{\text{WT}}(w_{(i)}^X)$, %
\begin{align*}
\alpha_{\mathrm{\comprehension}}^{\text{WT}}(w_{(i)}^X) &= \texttt{exact\_match}(\hat w_{(i)}^E, \Xi(w_{(i)}^X)) \\ %
&\hspace{-5em}\lor \texttt{inflection}(\hat w_{(i)}^E, \Xi(w_{(i)}^X)) \\
&\hspace{-5em}\lor \texttt{substring}(\hat w_{(i)}^E, \Xi(w_{(i)}^X))\\ %
&\hspace{-5em}\lor \texttt{inflection\_in\_substring}(\hat w_{(i)}^E, \Xi(w_{(i)}^X)) \\ %
&\hspace{-5em}\lor \texttt{synonym}(\hat w_{(i)}^{E}, \Xi(w_{(i)}^X)) \\ %
s_{\comprehension}^{\text{WT}}(w_{(i)}^X) &= \alpha_{\mathrm{\comprehension}}^{\text{WT}}(w_{(i)}^X) \in \{0,1\}
\end{align*}
where $\texttt{exact\_match}(\hat w_{(i)}^E, \Xi(w_{(i)}^X)) = 1$ if $\hat w_{(i)}^E$ matches with \textit{any} of the references in $\Xi(w_{(i)}^X)$ (analogously for other heuristics).

\paragraph{\generation} Given an English word $w_{(i)}^E$ and model prediction $\hat w_{(i)}^X$, we calculate binary accuracy for $w_{(i)}^E$ analogously to above, without considering  synonymy as we lack WordNets in our target LRLs. Recall that model scores are computed in terms of target language vocabulary, not English words. Suppose the word $w_{(m)}^X$ has $K=|\Xi(w_{(m)}^X)|$ English translations. We define $s_{\generation}^\text{WT}(w_{(m)}^X) \in [0,1]$ as %
\begin{align*}
s_{\generation}^{\text{WT}}(w_{(m)}^X) &= \frac{1}{K} \sum_{w_{(i)}^E \in \Xi(w_{(m)}^X)} \alpha_{\mathrm{\generation}}^{\text{WT}}(w_{(i)}^E)
\end{align*}

\subsection{Word Translation with Context}
Although a model may not understand or produce a word in isolation, it may do so given its natural context.  %
In this task, we provide additional context for the source-side language word in the form of a sentence containing it, and then prompt the LLM to perform word translation. %

This task requires monolingual data in the target language and in English for the \comprehension~ and \generation~ directions respectively.
We evaluate on all words in the available monolingual data that also have an entry in our bilingual lexicon. Note that the number of evaluated words may therefore differ by direction. 

\subsubsection{Scoring} 
The word to be translated $w_{(i)}$ may appear in several sentences. We define $C(w_{(i)})$ to be the number of times a word appears and $w_{(i,r)}$ 
to be the $r$th occurrence of word $w_{(i)}$.

\paragraph{\comprehension} 
We compute $\alpha_{\mathrm{\comprehension}}^{\text{WTWC}}(w_{(i, r)}^X)$ for a single occurrence similarly to $\alpha_{\mathrm{\comprehension}}^{\text{WT}}(w_{(i)}^X)$. We then average over occurrences to compute:
$$
s_{\comprehension}^{\text{WTWC}}(w_{(i)}^X) = \frac{1}{C(w_{(i)}^X)} \sum_{r=1}^{C(w_{(i)}^X)} \alpha_{\mathrm{\comprehension}}^{\text{WTWC}}(w_{(i, r)}^X)
$$

\paragraph{\generation} We evaluate \WordTranslationInContextAbbreviation~ \generation~ similarly to \WordTranslationAbbreviation~ \generation~ with
$$
\alpha_{\mathrm{\generation}}^{\text{WTWC}}(w_{(i,r)}^E) = \alpha_{\mathrm{\generation}}^{{\text{WT}}}(w_{(i)}^E) \in \{0, 1\}
$$

To account for $C(w_{(i)}^E)$ occurrences of $w_{(i)}^E$, we compute:
\begin{align*}
\beta_{\generation}^\text{WTWC}(w_{(i)}^E) &= \frac{1}{C(w_{(i)}^E)} \sum_{r=1}^{C(w_{(i)}^E)} \alpha_{\mathrm{\generation}}^{\text{WTWC}}(w_{(i,r)}^E)\\
s_{\generation}^{\text{WTWC}}(w_{(m)}^X) &= \frac{1}{K} \sum_{w_{(i)}^E \in \Xi(w_{(m)}^X)} \beta_{\generation}^{\text{WTWC}}(w_{(i)}^E)
\end{align*}

\subsection{Translation-Conditioned Language Modeling} 
\WordTranslationAbbreviation~ and \WordTranslationInContextAbbreviation~ prompt the model directly to comprehend or generate a word and utilize a binary accuracy metric for a given output. %
In \FillInTheBlankAbbreviation, we design a soft measure of the model's capability to do so given a sentence-level translation task.  %
We utilize parallel sentence pairs $t^X$-$t^E$ in target language $X$ and English, respectively. Given the entire source sentence and a partial translation up to the word of interest, we observe the generation probability of the correct word. %

Because this task deals with generation probabilities rather than observed outputs, we caution that the scores reported in each evaluation direction may not directly correspond to observed behavior. %
It may also not be comparable across models, as different models may have different generation distribution shapes. Similar to perplexity, this metric may be more useful in comparing various checkpoints of a single model.

\subsubsection{Scoring}
\paragraph{\generation}
We define the word of interest $w^X_{(i, r)}$ that appears at index $n$ in sentence $t^X$.
We provide the model with the complete sentence $t^E$ as well as the left context of $w^X_{(i, r)}$, denoted as $t_{<n}^X$. In $\alpha_{\mathrm{\generation}}^{\text{TCLM}}(w_{(i, r)}^X) \in [0,1]$, we observe the generation probability of $w^X_{(i, r)}$: %
\begin{align*}
    \alpha_{\mathrm{\generation}}^{\text{TCLM}}(w_{(i, r)}^X) &= P(w_{(i, r)}^X | t^E, t_{<n}^X) \\
    s_{\generation}^{\text{TCLM}}(w_{(i)}^X) &= \frac{1}{C(w_{(i)}^X)} \sum_{r=1}^{C(w_{(i)}^X)} \alpha_{\mathrm{\generation}}^{\text{TCLM}}(w_{(i, r)}^X)
    \label{eq:score_eng_to_x_tclm}
\end{align*}

Intuitively, this is a language-modeling-like task; however, pure language modeling without the source-side English sentence to guide the model has a higher entropy at every word, since the model may choose to continue with different concepts (not necessarily $w_{(i)}^X$) as reasonable continuations.
We use the sentence translation task to constrain the semantic scope of what the model might generate, thereby measuring the model's ability to generate a word broadly conditioned on its underlying concept.\footnote{We note that observing generation probabilities in this way is not a perfect measure of this ability. 
While the model knows the sentence-level semantics of the target language text as well as the left context up to the word of interest, it may still choose a different continuing formulation of the target-side sentence, leading to an unfairly low score.} 
Note that this evaluation does not require bilingual lexicons.

\paragraph{\comprehension} We now have $t^E$ on the output side, and are interested in evaluating the comprehension of various words $w_{(m)}^X$ in the source-side $t^X$ sentence. For every $w_{(i, r)}^E$ occurring at index $n$ of sentence $t^E$, we calculate
$$
\alpha_{\mathrm{\comprehension}}^{\text{TCLM}}(w_{(i, r)}^E) = P(w_{(i, r)}^E | t^X, t_{<n}^E) \in [0,1]
$$

The intuition is similar to the \generation~ case: we are interested in evaluating the model's ability to comprehend a word in a natural setting, and use the generation probability of its English equivalent given a restricted semantic scope. 
However, recall once again that \ProjectName~ scores are computed in terms of the vocabulary of the \emph{target language $X$}, not English. We therefore have the additional problem of finding the language $X$ word in $t^X$ that maps to or ``produced'' $w_{(i, r)}^E$. We use our existing lexicons in conjunction with statistical alignments with \ttt{FastAlign} \citep{dyer2013simple} to identify this mapping. We define an alignment as $\mathcal{A}(w_{(m)}^X) = \{w_{(i,r)}^E\}$ where $\mathcal{A}$ denotes alignments for sentence $t^X$-$t^E$. We define $\mathcal{F}$ as a union of $\Xi(w_{(m)}^X)$ and $\mathcal{A}(w_{(m)}^X)$, prioritizing the former. %
For every $w_{(m)}^X \in t^X$, we calculate:
\begin{align*}
    \beta_{\comprehension}^{\text{TCLM}}(w_{(i)}^E) &= \frac{1}{C(w_{(i)}^E)} \sum_{r=1}^{C(w_{(i)}^X)}\alpha_{\comprehension}^{\text{TCLM}}(w_{(i,r)}^E) \\
    s_{\comprehension}^{\text{TCLM}}(w_{(m)}^X) &= \frac{1}{\mathcal{|F|}}\sum_{w_{(i)}^E \in \mathcal{F}} \beta_{\comprehension}^\text{TCLM} (w_{(i)}^E)
\end{align*}

\subsection{Bag-of-Words Machine Translation}
Given a sequence-level machine translation task, metrics such as \textsc{BLEU} \citep{papineni2002bleu} and \textsc{chrF} \citep{popovic2015chrf}  measure translation quality by assessing the exact match n-gram or character-gram overlap between model outputs and reference translations. %
Given our lexical focus, we instead formulate a coarser evaluation metric. %
Given a sentence-level MT task, we are interested in evaluating whether the target language words were correctly produced (\generation) or translated correctly to English equivalents (\comprehension), regardless of the syntax of the output or the appropriateness of the morphological form of the word. 

\subsubsection{Scoring}
\paragraph{\generation} Given a parallel sentence pair $t^X$-$t^E$, we prompt $M_{(\kappa)}$ to translate $t^E$ to target language $X$. For every $w_{(i)}^{X} \in t^X$, we check whether the predicted sentence $\hat t^X$ contains $w_{(i)}^{X}$. %
We calculate:
\begin{align*}
\alpha_{\mathrm{\generation}}^{\text{BOW MT}}(w_{(i,r)}^X) &= \texttt{exact\_match}(\hat t^X, w_{(i)}^X) \\
&\hspace{-5em} \lor \texttt{inflection}(\hat t^X, w_{(i)}^X) \\
s_\generation^{\text{BOW MT}}(w_{(i)}^X) &= \frac{1}{C(w_{(i)}^X)} \sum_{r=1}^{C(w_{(i)}^X)} \alpha_{\mathrm{\generation}}^{\text{BOW MT}}(w_{(i,r)}^X)
\end{align*}
\paragraph{\comprehension} Given $t^X$-$t^E$, we prompt $M_{(\kappa)}$ to translate $t^X$ into English. We check whether the predicted sentence $\hat t^E$ contains $w_{(i)}^E \in t^E$. 
\begin{align*}
\alpha_{\mathrm{\comprehension}}^{\text{BOW MT}}(w_{(i, r)}^E) &= \texttt{exact\_match}(\hat t^E, w_{(i)}^E) \notag \\
    &\hspace{-5em} \lor \texttt{inflection} (\hat t^E, w_{(i)}^E) \lor \texttt{synonym} (\hat t^E, w_{(i)}^E) \\ 
\end{align*}
Similarly as in \FillInTheBlankAbbreviation, we generate the English alignments $\mathcal{F}$ for $w_{(m)}^X$ and compute its score:
\begin{align*}
\beta_{\comprehension}^{\text{BOW MT}}(w_{(i)}^E) &= \frac{1}{C(w_{(i)}^E)} \sum_{r=1}^{C(w_{(i)}^E)} \alpha_{\comprehension}^{\text{BOW MT}}(w_{(i,r)}^E) \\
s_{\comprehension}^\text{BOW MT}(w_{(m)}^X) &= \frac{1}{|\mathcal{F}|} \sum_{w_{(i)}^E \in \mathcal{F}} \beta_{\comprehension}^\text{BOW MT}(w_{(i)}^E)
\end{align*}

\begin{figure*}[!ht]
    \centering
    \includegraphics[width=1\linewidth]{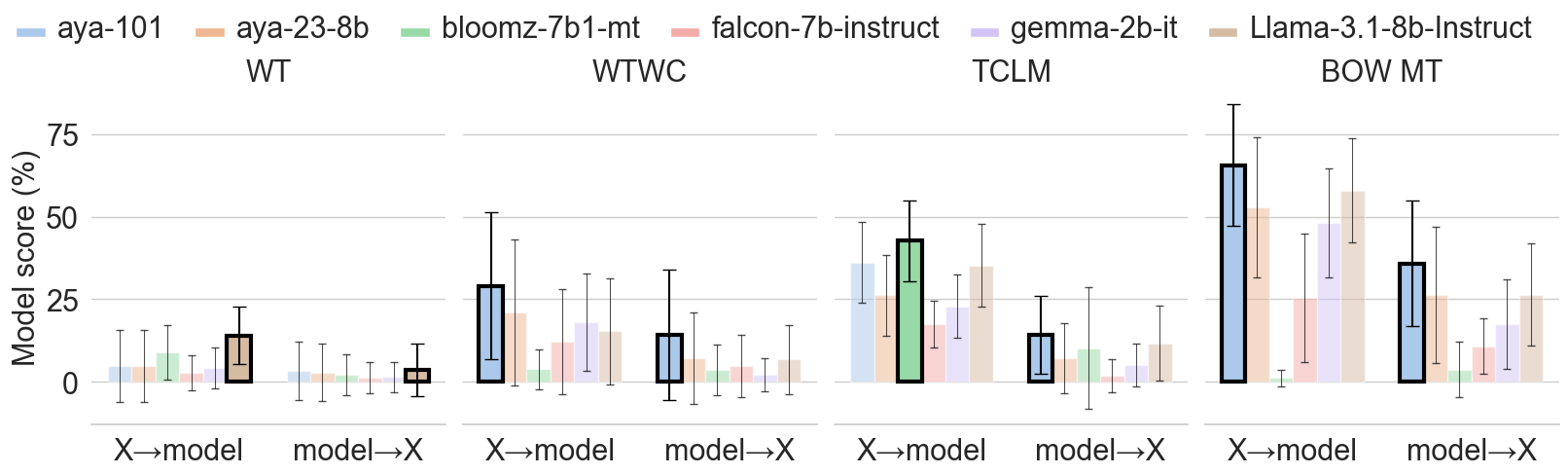}
    \caption{Model scores across subtasks, with std. deviation over languages. Best performing model is highlighted.} %
    \label{fig:average_model_performance_across_tasks}
\end{figure*}
\begin{figure*}[!h]
    \centering
    \includegraphics[width=1\linewidth]{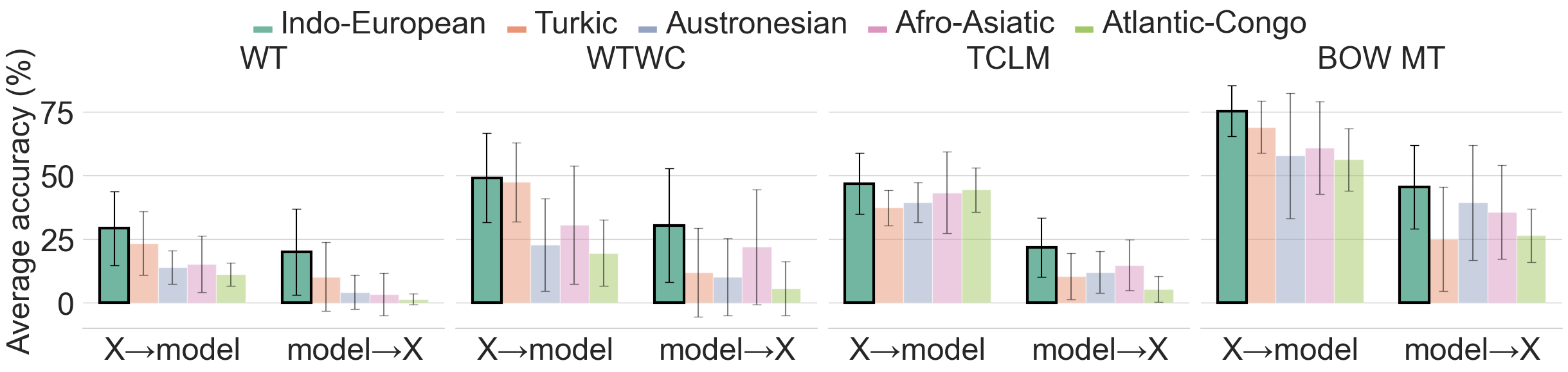}
    \caption{We compute the score of a language family as the average of its constituent languages, with the best-performing language family highlighted. Error bars represent the standard deviation within the language family. The Indo-European family has consistently higher scores than other families.}
    \label{fig:performance_across_lang_families}
\end{figure*}

\section{Data Sources and Languages}

\paragraph{Data sources}
\begin{table}[h]
    \centering
    \small
    \begin{tabular}{c | p{2.4cm} | p{2.4cm}}
        \toprule
        \textbf{Task} & \textbf{\comprehension} & \textbf{\generation} \\
        \midrule
        \WordTranslationAbbreviation & lexicons & lexicons \\
        \midrule
        \WordTranslationInContextAbbreviation & lexicons, monolingual datasets & lexicons, monolingual datasets \\
        \midrule
        \FillInTheBlankAbbreviation & lexicons, bitext & bitext \\
        \midrule
        \BagOfWordsMachineTranslationAbbreviation & lexicons, bitext & bitext \\
        \bottomrule
    \end{tabular}
    \caption{Data type required in each task}
    \label{tab:datasets_required_for_each_task}
\end{table}

\autoref{tab:datasets_required_for_each_task} lists the types of data required for each task, as per the task description above. 
We use \textbf{lexicons} created by amalgamating GATITOS \citep{jones-etal-2023-gatitos}, Intercontinental Dictionary Series \citep{hans_jorg_bibiko_2023_7701635}, and PanLex \citep{kamholz-etal-2014-panlex} data. For a given word, we used translations from the first two if available, and fallback to PanLex.
See \autoref{sec:lexicons} for language coverage of lexicons. %
We use \textbf{monolingual data} from GLOTLID \citep{Kargaran_2023} which covers 1665 languages,  %
and \textbf{parallel data} from FLORES+ \citep{nllb-24}, which covers 212 languages.
We discard languages with fewer than 100 entries in the target language lexicon. In \WordTranslationInContextAbbreviation, \FillInTheBlankAbbreviation, and \BagOfWordsMachineTranslationAbbreviation~ we discard languages where our lexicons cover less than $100$ unique words from monolingual or parallel data.

\paragraph{Languages}
See \autoref{sec:languages} for details concerning the distribution of languages over language families as covered by each task, geographic spread, and code conventions used.

\section{Experimental Setup}

We evaluated six multilingual open-source models: \ttt{\ayaOne} \citep{ustun2024ayamodelinstructionfinetuned}, \ttt{\ayaTwo} \citep{aryabumi2024aya23openweight}, \ttt{\bloom} \citep{muennighoff2023crosslingualgeneralizationmultitaskfinetuning}, \ttt{\falcon} \citep{almazrouei2023falconseriesopenlanguage}, \ttt{\gemma} \citep{gemmateam2024gemma2improvingopen}, and \ttt{\llama} \citep{grattafiori2024llama}.
We list key characteristics of these models in \autoref{tab:model_overview}.
See \autoref{sec:prompt_exploration} for prompts used per subtask and \autoref{sec:evaluation_details} for details on GPU hours required to run each subtask.

Given the number of languages and size of dataset, we present a \texttt{lite} version of the benchmark, which uses a subset of the data as follows. 
We cap the number of vocabulary entries per language for \WordTranslationAbbreviation~ and \WordTranslationInContextAbbreviation~ \comprehension~ at 300. 
We use 30\% of the available data for \FillInTheBlankAbbreviation~ and \BagOfWordsMachineTranslationAbbreviation.
All reported language scores are computed over a minimum of 100 words per language.

\section{Results and Discussion} \label{sec:results_and_discussion}

See the performance of tested models on all $8$ subtasks in \autoref{fig:average_model_performance_across_tasks}. See detailed results in \autoref{sec:results_in_detail}, including the language score distribution per subtask and model as well as sampled language scores. %
Broadly, we observe that models have significant room for improvement; i.e. \textbf{our benchmark is a challenging measure of multilingual performance}.

We train a decision tree to predict language scores per task based on a series of features, including model, language resourcedness, script, language family, and others. We find the top features that determine task performance for a given language are evaluation direction, whether the language is supported by the model, and resource level of the language (see \autoref{subsec:feature_importance} for decision trees and ranked feature importances).
We discuss these features in more detail below.

\paragraph{Evaluation direction}
Models evaluated in the \comprehension~ direction exhibit higher scores than in the \generation~ direction, i.e. even if a model can comprehend a word in an LRL, it might not be capable of generating it. This finding is consistent with previous literature that finds a considerable gap between NLU and NLG, or the out-of-X direction and the into-X direction in MT \citep{belinkov2017neural, kandimalla2022improving}.

\begin{figure}[!h]
    \centering
    \vspace{-1em}
    \includegraphics[width=\linewidth]{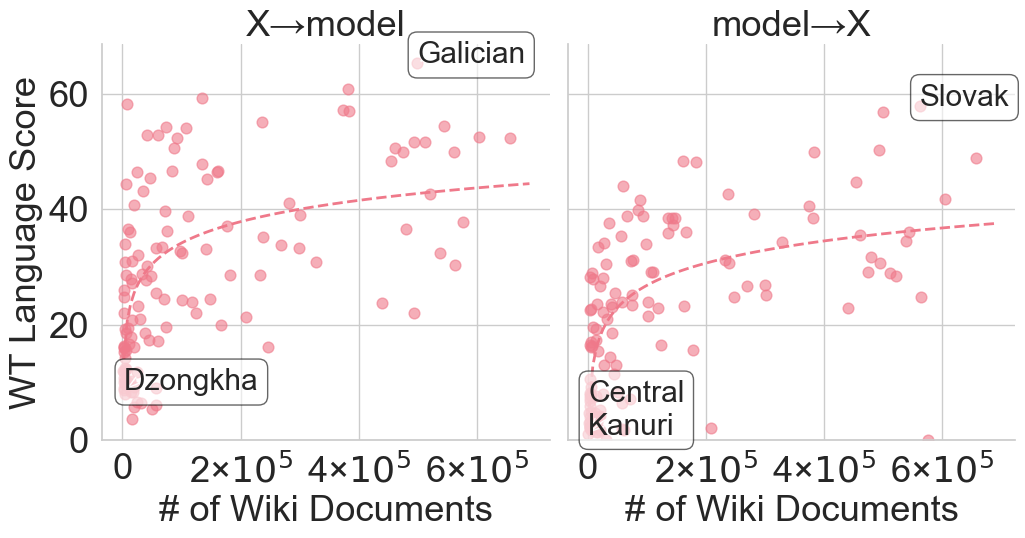}
    \caption{Comparison of the number of Wikipedia documents\textemdash a proxy for resource level\textemdash and language performance for the task \WordTranslationAbbreviation. See \autoref{subsec:resourceness} for other tasks.} %
    \label{fig:wt_performance_versus_resource_level}
\end{figure}

\begin{figure}[!h]
    \centering
    \includegraphics[width=\linewidth]{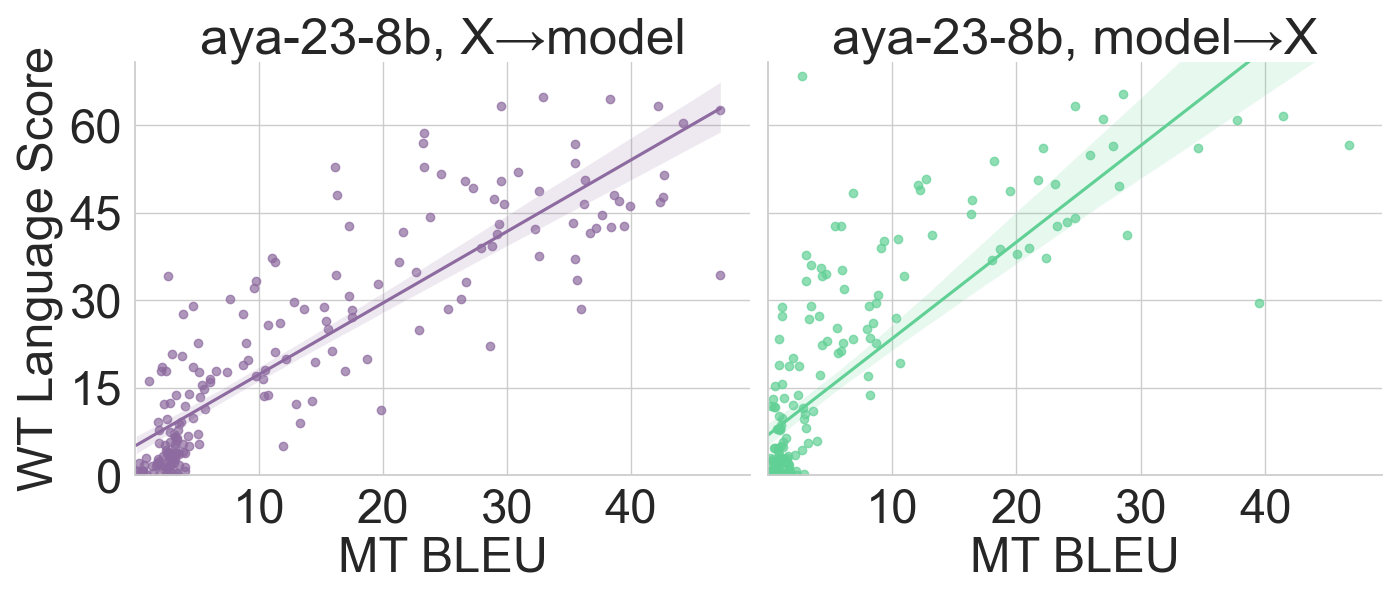}
    \caption{\WordTranslationAbbreviation~ scores are strongly correlated with sentence-level MT \textsc{BLEU} scores.} %
    \label{fig:machine_versus_word_translation}
\end{figure}

\paragraph{Language family and resource} In \autoref{fig:performance_across_lang_families}, we draw attention to the performance gap between Indo-European languages and underrepresented Austronesian and Atlantic-Congo languages. %

Naturally, there is a lot of variation between model performance on languages within a single family, depending on other potential factors such as the resourcedness of the language and whether it is supported by the model. In \autoref{fig:wt_performance_versus_resource_level}, we show the relationship between resource level and \WordTranslationAbbreviation~ performance. This is roughly logarithmic, with the bulk of LRLs performing significantly worse than HRLs, large improvements for mid-resource languages, and gains saturating for HRLs. %
In sum, \textbf{we highlight the scope of improvement for SOTA models on underrepresented language families and low-resource languages}.

\paragraph{Model} \ayaOne~ achieves the highest average score on five of the eight subtasks. Compared to other models, \ayaOne~ is unique in that it employs an encoder-decoder architecture, is larger (13B parameters), and instruction-tuned on 101 languages. (See \autoref{tab:model_overview}). These qualities may contribute to its performance.

\paragraph{Task} Note that while \WordTranslationAbbreviation, \WordTranslationInContextAbbreviation, and \BagOfWordsMachineTranslationAbbreviation~ all report accuracy metrics over a vocabulary set, model scores are not directly comparable across tasks as they are computed over different vocabularies, as per resource requirements for each task. That being said, we observe generally higher scores for \WordTranslationInContextAbbreviation~ than \WordTranslationAbbreviation~ in \autoref{fig:average_model_performance_across_tasks}. 
This indicates that models are able to utilize and benefit from the additional context provided in the former.

We also see that models generally show higher scores for \BagOfWordsMachineTranslationAbbreviation~ than for \WordTranslationAbbreviation~ and \WordTranslationInContextAbbreviation. 
\BagOfWordsMachineTranslationAbbreviation~ uses a sentence-level machine translation setup, which instruction-tuned models may be more familiar with as opposed to direct prompts concerning word meaning as used in \WordTranslationAbbreviation~ and \WordTranslationInContextAbbreviation. 
\BagOfWordsMachineTranslationAbbreviation~ also allows the model to generate the previous context of the word of interest in the output translation, potentially priming the model better in terms of semantic context as well as language of generation.

As discussed in \autoref{sec:dataset_description}, \FillInTheBlankAbbreviation~ is less directly interpretable and comparable across models than the other tasks and is better employed during model development. \textbf{By including subtasks of different difficulties and settings, our benchmark allows for various perspectives and a nuanced understanding of lexical competence.}

\paragraph{Correlation with MT}
While machine translation is a good measure of natural language understanding \citep{iyer2023towards}, sentence-level translation datasets are expensive to create and curate. 
In \autoref{fig:machine_versus_word_translation}, we demonstrate that there is a strong linear correlation between BLEU scores on machine translation performance with FLORES+ and scores from \WordTranslationAbbreviation, for available languages in FLORES+ (0.873 and 0.769 in the \comprehension~ and \generation~ evaluation direction respectively).
Given that \WordTranslationAbbreviation~ covers $2700+$ languages as opposed to the $212$ covered by FLORES+, \textbf{our benchmark can provide a cheap proxy in the absence of machine translation data}. %

\section{Conclusion}

We introduce \ProjectName, a massively multilingual benchmark testing lexical competence, that draws on existing available resources such as lexicons, monolingual data, and bitext. 
\ProjectName~ consists of $8$ subtasks that provide various perspectives on lexical comprehension and generation skills.
We evaluate SOTA models on our benchmark and find that these have a long way to go for low-resource languages.
With this work, we hope to promote the massively multilingual evaluation of LLMs as one step towards addressing language inequity in NLP.

\section{Limitations}

\paragraph{Benchmark coverage and quality} 
While \ProjectName~ covers 2700+ languages, this is restricted to the relatively simple word translation task; other tasks have more typical coverage compared to existing benchmarks. Further, \ProjectName~ is heavily reliant on public lexicons, specifically PanLex, which is known to be noisy for many languages. 
We note that while \ProjectName~ currently relies heavily on PanLex, GLOTLID, and FLORES+, it can integrate other lexicons and monolingual and parallel datasets in the future. 
Notably, the  BOUQuET dataset \citep{andrews-etal-2025-bouquet,omnilingual2026}, covering 275 languages, was released after the completion of this work, and constitutes an additional high-quality source of parallel data.

\paragraph{Lexical coverage, sense disambiguation, and synonymy} 
Our benchmark is limited by the available annotations in the lexicons we work with. This results in a number of shortcomings and avenues for future improvement.
Several languages may only have a few hundred entries in available lexicons. Further, models may output valid variants or synonyms that are not documented in our lexicons, potentially resulting in false negatives in \WordTranslationAbbreviation. 
Our lexicons also do not annotate word sense. This limitation may become problematic, e.g. in \WordTranslationInContextAbbreviation~ where only a particular word sense should be marked correct given a sentence.

\paragraph{Morphological, syntactic, and complex semantic skills are out of scope.} Our benchmark focuses on evaluating lexical understanding in models. However, basic skills in a language also include understanding and producing appropriate morphological forms and appropriate word orders for utterances. 
Although these are important dimensions of the evaluation, we currently lack resources in the target languages to evaluate these skills in our benchmark. 

We hope that our experiments and benchmark motivate the further collection and refinement of lexicons, as well as other such resources in low-resource languages. In doing so, \ProjectName~ can enable richer evaluations of the basic linguistic skills of LLMs on a massively multilingual scale.

\section*{Ethics Statement}

We do not expect any negative ethical consequences of this work, which presents a benchmark for the multilingual evaluation of large language models. 
We use publicly available datasets to design our benchmark, and provide results on open-source models. Our benchmark and Python package are in accordance with the licenses of each constituent dataset (see \autoref{sec:licensing}) and include functionalities for downloading the data as well as running evaluations. 
This work used LLMs for coding assistance only.

\section{Acknowledgments}

We would like to thank Drs. David Yarowsky and Karen Livescu for helpful discussions and feedback on this paper. We also thank the anonymous reviewers for their feedback.

\bibliography{acl}

\appendix
\appendix
\newpage
\onecolumn
\section{Prior Work} \label{sec:prior_work}
\begin{table*}[!hbtp]
    \centering
    \small
    \resizebox{0.47\linewidth}{!}{
    \begin{tabular}{l | l | l}
        \toprule
        \textbf{Benchmark} & \textbf{Task} & \makecell[l]{\textbf{No. of} \\ \textbf{Langages}} \\
        \midrule
        \multicolumn{3}{l}{\textbf{SOTA Benchmarks}} \\
        \midrule
        \makecell[l]{FLORES-200\\\citep{nllb-24}} & Translation & 212 \\ \midrule
        \makecell[l]{BELEBELE\\\citep{bandarkar-etal-2024-belebele}} & \makecell[l]{Reading\\Comprehension} & 122 \\ \midrule
        \makecell[l]{Aya Evaluation Suite\\\citep{singh2024aya}} & \makecell[l]{Instruction\\Following} & 101 \\ \midrule
        MEGA \citep{ahuja2023mega} & \makecell[l]{Generation, \\Classification} & 70 \\ \midrule
        XL-Sum \citep{hasan-etal-2021-xl} & Summarization & 43 \\ \midrule
        MaXIFE \citep{liu2025maxife} & \makecell[l]{Instruction\\Following} & 23 \\ \midrule
        \makecell[l]{Aya Expanse, m-Arena Hard \\ \citep{dang2024ayaexpansecombiningresearch}} & \makecell[l]{Instruction\\Following} & 23 \\ \midrule
        \makecell[l]{WikiLingua \\\citep{ladhak-etal-2020-wikilingua}} & Summarization & 18 \\ \midrule
        \makecell[l]{MMMLU \\ \citep{hendrycks2020measuring}} & Reasoning & 14 \\ \midrule
        XNLI \citep{conneau2018xnlievaluatingcrosslingualsentence} & Inference & 14 \\ \midrule
        XCOPA \citep{ponti2020xcopa} & Classification & 11 \\ \midrule
        \makecell[l]{XStoryCloze \\ \citep{lin_et_al_x_storycloze_2021}} & Reasoning & 11 \\ \midrule
        TyDiQA \citep{clark-etal-2020-tydi} & \makecell[l]{Question\\Answering}& 11 \\ \midrule
        GSM8K \citep{cobbe2021gsm8k} & \makecell[l]{Mathematical\\Reasoning} & 10 \\ \midrule
        \makecell[l]{M3Exam\\\citep{zhang2023m3exam}} & \makecell[l]{Question\\Answering} & 9 \\ \midrule
        PAWS-X \citep{yang2019paws} & \makecell[l]{Paraphrase\\Identification} & 6 \\ \midrule
        MLQA \citep{lewis-etal-2020-mlqa} & \makecell[l]{Question\\Answering} & 7 \\ \midrule
        \makecell[l]{XWinograd\\\citep{muennighoff2023crosslingualgeneralizationmultitaskfinetuning}} & \makecell[l]{Coreference\\Resolution} & 6 \\ \midrule %
        Dolly \citep{DatabricksBlog2023DollyV2} & \makecell[l]{Instruction\\Following} & 3 \\ \midrule
        $\infty$Bench \citep{zhang2024inftybenchextendinglongcontext} & \makecell[l]{Long Context\\Reasoning} & 2 \\ 
        \midrule
        \multicolumn{3}{l}{\textbf{Lexical Understanding}} \\
        \midrule
        \makecell[l]{MuCoW\\\citep{raganato-etal-2019-mucow}} & \makecell[l]{Lexical\\Substitution} & 12 \\ \midrule
        \makecell[l]{ContraWSD \\\citep{rios-gonzales-etal-2017-improving}} & \makecell[l]{Lexical\\Substitution} & 3 \\ \midrule
        \makecell[l]{Cross-lingual Lexical\\Substitution Task\\ \citep{mihalcea-etal-2010-semeval}} & \makecell[l]{Lexical\\Substitution} & 2  \\ \midrule
        \makecell[l]{TOEFL, StuVoc, LexTale \\\citep{martinez_et_al_vocab_tests_2024}} & \makecell[l]{Lexical\\Substitution} & 2 \\ \midrule
        \makecell[l]{Word Sense Disambiguation\\Test Suite \citep{rios-etal-2018-word}} & \makecell[l]{Lexical\\Substitution} & 2 \\ \midrule
        \makecell[l]{Danish Semantic Reasoning\\Benchmark\\\citep{pedersen-etal-2024-towards}} & \makecell[l]{Lexical\\Substitution} & 1 \\ \midrule
        \textbf{\ProjectName} & \makecell[l]{\textbf{Lexical}\\ \textbf{Comprehension}\\\textbf{and Generation}} & \textbf{2746} \\
        \bottomrule
    \end{tabular}}
    \caption{Language coverage across text benchmarks that evaluate multilingual NLU and NLG capabilities.}
    \label{tab:prior_multiling_benchmark_coverage}
\end{table*}

\autoref{tab:prior_multiling_benchmark_coverage} lists SOTA multilingual benchmarks as well as past work performed in lexical substitution, a task that is closest with our current work. The listed benchmarks exhibit limited language coverage.

\section{Languages} \label{sec:languages}
\subsection{Geographic Spread}

\begin{figure*}
    \centering
    \includegraphics[width=1\linewidth]{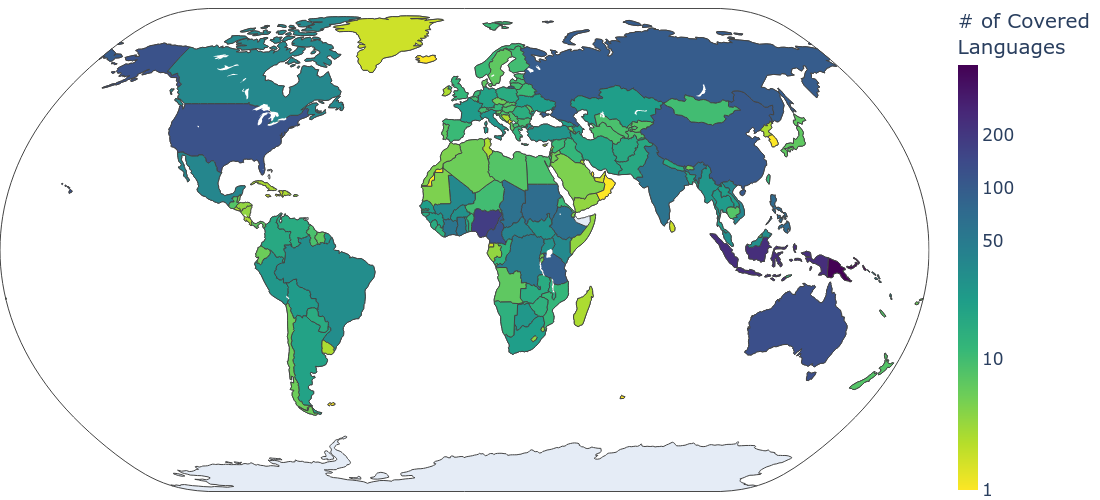}
    \caption{Drawing on Glottolog data \citep{glottolog52}, the choropleth map above illustrates the geographic distribution of the languages covered in at least one task in \ProjectName. Specifically, we note the country of origin. Countries with the highest number of languages include: Papua New Guinea where 498 languages originate,  Indonesia 240, and Nigeria 182.}
    \label{fig:geographic_spread_of_languages}
\end{figure*}

\autoref{fig:geographic_spread_of_languages} notes the country of origin of languages represented in at least one task. %
In observing the geographic spread of this task, we see that we attain coverage in all world countries.

\subsection{Language Families} \label{subsec:language_families}
In \autoref{tab:distribution_of_language_families_across_tasks_table1}, \autoref{tab:distribution_of_language_families_across_tasks_table2}, and \autoref{tab:distribution_of_language_families_across_tasks_table3}, we report the number of languages in Glottolog families in each of the four tasks. 
\begin{table}[!hbtp]
    \centering
    \small
    \resizebox{0.45\textwidth}{!}{%
        \begin{tabular}{l | r | r | r | r}
            \toprule
            \textbf{Language Family} & \textbf{\texttt{\WordTranslationAbbreviation}} & \textbf{\texttt{\WordTranslationInContextAbbreviation}} & \textbf{\texttt{\FillInTheBlankAbbreviation}} & \textbf{\texttt{\BagOfWordsMachineTranslationAbbreviation}} \\
            \midrule
            Atlantic-Congo & 483 & 85 & 33 & 33 \\
            Austronesian & 483 & 103 & 21 & 21 \\
            \makecell[l]{Nuclear Trans\\New Guinea} & 225 & 15 & 0 & 0 \\
            Indo-European & 184 & 123 & 73 & 73 \\
            Afro-Asiatic & 134 & 21 & 19 & 19 \\
            Pama-Nyungan & 71 & 1 & 0 & 0 \\
            Tai-Kadai & 38 & 2 & 3 & 3 \\
            Sino-Tibetan & 38 & 11 & 9 & 9 \\
            Mande & 32 & 3 & 2 & 2 \\
            Nakh-Daghestanian & 29 & 10 & 0 & 0 \\
            Uralic & 28 & 18 & 4 & 4 \\
            Nuclear Torricelli & 27 & 0 & 0 & 0 \\
            Sepik & 27 & 2 & 0 & 0 \\
            Austroasiatic & 26 & 5 & 3 & 3 \\
            \makecell[l]{Athabaskan-Eyak-\\Tlingit} & 26 & 4 & 0 & 0 \\
            Turkic & 25 & 27 & 14 & 14 \\
            Artificial Language & 18 & 10 & 2 & 2 \\
            Central Sudanic & 16 & 0 & 0 & 0 \\
            Quechuan & 16 & 11 & 1 & 1 \\
            Uto-Aztecan & 16 & 2 & 0 & 0 \\
            Dogon & 16 & 0 & 0 & 0 \\
            Timor-Alor-Pantar & 16 & 1 & 0 & 0 \\
            Nilotic & 15 & 6 & 3 & 3 \\
            Algic & 15 & 1 & 0 & 0 \\
            Ta-Ne-Omotic & 14 & 0 & 0 & 0 \\
            Hmong-Mien & 14 & 1 & 0 & 0 \\
            Otomanguean & 13 & 1 & 0 & 0 \\
            Kru & 13 & 0 & 0 & 0 \\
            Angan & 12 & 0 & 0 & 0 \\
            Arawakan & 11 & 1 & 0 & 0 \\
            Khoe-Kwadi & 10 & 1 & 0 & 0 \\
            Dravidian & 10 & 1 & 4 & 4 \\
            Pano-Tacanan & 10 & 1 & 0 & 0 \\
            Surmic & 10 & 0 & 0 & 0 \\
            Heibanic & 10 & 0 & 0 & 0 \\
            Nyulnyulan & 9 & 0 & 0 & 0 \\
            Anim & 9 & 0 & 0 & 0 \\
            Mayan & 9 & 5 & 0 & 0 \\
            Gunwinyguan & 8 & 0 & 0 & 0 \\
            Tupian & 8 & 3 & 1 & 1 \\
            Yam & 8 & 0 & 0 & 0 \\
            Dagan & 8 & 1 & 0 & 0 \\
            Cariban & 8 & 3 & 0 & 0 \\
            Ramu & 8 & 1 & 0 & 0 \\
            South Bird's Head & 7 & 0 & 0 & 0 \\
            Nubian & 7 & 0 & 0 & 0 \\
            Bosavi & 7 & 0 & 0 & 0 \\
            Pomoan & 7 & 0 & 0 & 0 \\
            Kadugli-Krongo & 6 & 0 & 0 & 0 \\
            Mailuan & 6 & 0 & 0 & 0 \\
            Ndu & 6 & 0 & 0 & 0 \\
            Saharan & 6 & 0 & 2 & 2 \\
            Siouan & 6 & 0 & 0 & 0 \\
            Left May & 6 & 0 & 0 & 0 \\
            Koiarian & 6 & 2 & 0 & 0 \\
            Japonic & 6 & 1 & 1 & 1 \\
            Kiwaian & 6 & 0 & 0 & 0 \\
            Tungusic & 6 & 0 & 0 & 0 \\
            Lower Sepik & 5 & 0 & 0 & 0 \\
            Eleman & 5 & 1 & 0 & 0 \\
            Cochimi-Yuman & 5 & 0 & 0 & 0 \\
            Narrow Talodi & 5 & 0 & 0 & 0 \\
            South Bougainville & 5 & 0 & 0 & 0 \\
            Yeniseian & 5 & 0 & 0 & 0 \\
            \bottomrule
        \end{tabular}%
    }
    \caption{Distribution of languages in Glottolog language families across all tasks.}
    \label{tab:distribution_of_language_families_across_tasks_table1}
\end{table}

\begin{table}[!hbtp]
    \centering
    \small
    \resizebox{0.45\textwidth}{!}{%
        \begin{tabular}{l | r | r | r | r}
            \toprule
            \textbf{Language Family} & \textbf{\texttt{\WordTranslationAbbreviation}} & \textbf{\texttt{\WordTranslationInContextAbbreviation}} & \textbf{\texttt{\FillInTheBlankAbbreviation}} & \textbf{\texttt{\BagOfWordsMachineTranslationAbbreviation}} \\
            \midrule
            Muskogean & 5 & 1 & 0 & 0 \\
            Miwok-Costanoan & 5 & 0 & 0 & 0 \\
            Eskimo-Aleut & 5 & 2 & 0 & 0 \\
            East Strickland & 5 & 0 & 0 & 0 \\
            Salishan & 5 & 0 & 0 & 0 \\
            Yareban & 5 & 1 & 0 & 0 \\
            Mataguayan & 5 & 1 & 0 & 0 \\
            Suki-Gogodala & 4 & 0 & 0 & 0 \\
            Lengua-Mascoy & 4 & 0 & 0 & 0 \\
            Eastern Trans-Fly & 4 & 0 & 0 & 0 \\
            Kartvelian & 4 & 2 & 1 & 1 \\
            Abkhaz-Adyge & 4 & 3 & 0 & 0 \\
            Koman & 4 & 0 & 0 & 0 \\
            Ijoid & 4 & 0 & 0 & 0 \\
            Mangarrayi-Maran & 4 & 0 & 0 & 0 \\
            Eastern Jebel & 4 & 0 & 0 & 0 \\
            Songhay & 4 & 2 & 0 & 0 \\
            Maban & 4 & 0 & 0 & 0 \\
            Tuu & 4 & 0 & 0 & 0 \\
            Iroquoian & 4 & 1 & 0 & 0 \\
            Dajuic & 4 & 0 & 0 & 0 \\
            Guaicuruan & 4 & 0 & 0 & 0 \\
            Chumashan & 4 & 0 & 0 & 0 \\
            Mirndi & 4 & 0 & 0 & 0 \\
            North Bougainville & 4 & 0 & 0 & 0 \\
            Tangkic & 3 & 0 & 0 & 0 \\
            South Omotic & 3 & 0 & 0 & 0 \\
            Kuliak & 3 & 0 & 0 & 0 \\
            Kwalean & 3 & 0 & 0 & 0 \\
            Kxa & 3 & 0 & 0 & 0 \\
            Kamula-Elevala & 3 & 0 & 0 & 0 \\
            Kolopom & 3 & 0 & 0 & 0 \\
            Chibchan & 3 & 1 & 0 & 0 \\
            Iwaidjan Proper & 3 & 0 & 0 & 0 \\
            Bookkeeping & 3 & 0 & 0 & 0 \\
            Mongolic-Khitan & 3 & 3 & 1 & 1 \\
            West Bomberai & 3 & 0 & 0 & 0 \\
            Chocoan & 3 & 0 & 0 & 0 \\
            Jarrakan & 3 & 0 & 0 & 0 \\
            Maningrida & 3 & 1 & 0 & 0 \\
            Nuclear-Macro-Je & 3 & 1 & 0 & 0 \\
            Dizoid & 3 & 0 & 0 & 0 \\
            Tucanoan & 3 & 0 & 0 & 0 \\
            Walioic & 3 & 0 & 0 & 0 \\
            Tamaic & 3 & 0 & 0 & 0 \\
            Konda-Yahadian & 2 & 0 & 0 & 0 \\
            Rashad & 2 & 0 & 0 & 0 \\
            Keram & 2 & 0 & 0 & 0 \\
            Haida & 2 & 0 & 0 & 0 \\
            Mixe-Zoque & 2 & 0 & 0 & 0 \\
            Yanomamic & 2 & 0 & 0 & 0 \\
            Bogia & 2 & 0 & 0 & 0 \\
            Caddoan & 2 & 0 & 0 & 0 \\
            Kunimaipan & 2 & 0 & 0 & 0 \\
            Pahoturi & 2 & 0 & 0 & 0 \\
            Baibai-Fas & 2 & 0 & 0 & 0 \\
            Kayagaric & 2 & 0 & 0 & 0 \\
            Sign Language & 2 & 0 & 0 & 0 \\
            Katla-Tima & 2 & 0 & 0 & 0 \\
            Yangmanic & 2 & 0 & 0 & 0 \\
            Kresh-Aja & 2 & 0 & 0 & 0 \\
            Piawi & 2 & 0 & 0 & 0 \\
            Kwomtari-Nai & 2 & 0 & 0 & 0 \\
            Arafundi & 2 & 0 & 0 & 0 \\
            \bottomrule
        \end{tabular}%
    }
    \caption{Distribution of languages in Glottolog language families across all tasks.}
    \label{tab:distribution_of_language_families_across_tasks_table2}
\end{table}

\begin{table}[!hbtp]
    \centering
    \small
    \resizebox{0.45\textwidth}{!}{%
        \begin{tabular}{l | r | r | r | r}
            \toprule
            \textbf{Language Family} & \textbf{\texttt{\WordTranslationAbbreviation}} & \textbf{\texttt{\WordTranslationInContextAbbreviation}} & \textbf{\texttt{\FillInTheBlankAbbreviation}} & \textbf{\texttt{\BagOfWordsMachineTranslationAbbreviation}} \\
            \midrule
            Somahai & 2 & 0 & 0 & 0 \\
            Bunaban & 2 & 0 & 0 & 0 \\
            Kaure-Kosare & 2 & 0 & 0 & 0 \\
            Bayono-Awbono & 2 & 0 & 0 & 0 \\
            Giimbiyu & 2 & 0 & 0 & 0 \\
            Bulaka River & 2 & 0 & 0 & 0 \\
            Teberan & 2 & 1 & 0 & 0 \\
            Mombum-Koneraw & 2 & 0 & 0 & 0 \\
            Worrorran & 2 & 0 & 0 & 0 \\
            Manubaran & 2 & 0 & 0 & 0 \\
            Chonan & 2 & 0 & 0 & 0 \\
            Barbacoan & 2 & 0 & 0 & 0 \\
            Amto-Musan & 2 & 0 & 0 & 0 \\
            Turama-Kikori & 2 & 0 & 0 & 0 \\
            Maiduan & 1 & 0 & 0 & 0 \\
            Chicham & 1 & 0 & 0 & 0 \\
            Koreanic & 1 & 1 & 1 & 1 \\
            Lakes Plain & 1 & 0 & 0 & 0 \\
            Gumuz & 1 & 0 & 0 & 0 \\
            Aymaran & 1 & 1 & 1 & 1 \\
            Temeinic & 1 & 0 & 0 & 0 \\
            Chukotko-Kamchatkan & 1 & 0 & 0 & 0 \\
            Kawesqar & 1 & 0 & 0 & 0 \\
            Huitotoan & 1 & 0 & 0 & 0 \\
            Misumalpan & 1 & 0 & 0 & 0 \\
            Kiowa-Tanoan & 1 & 0 & 0 & 0 \\
            Wakashan & 1 & 0 & 0 & 0 \\
            Arawan & 1 & 1 & 0 & 0 \\
            Garrwan & 1 & 0 & 0 & 0 \\
            Tarascan & 1 & 1 & 0 & 0 \\
            Chinookan & 1 & 0 & 0 & 0 \\
            Saliban & 1 & 0 & 0 & 0 \\
            East Kutubu & 1 & 0 & 0 & 0 \\
            Totonacan & 1 & 0 & 0 & 0 \\
            Sahaptian & 1 & 0 & 0 & 0 \\
            Zamucoan & 1 & 0 & 0 & 0 \\
            Tsimshian & 1 & 0 & 0 & 0 \\
            Ainu & 1 & 0 & 0 & 0 \\
            Tequistlatecan & 1 & 0 & 0 & 0 \\
            Great Andamanese & 1 & 0 & 0 & 0 \\
            Peba-Yagua & 1 & 0 & 0 & 0 \\
            Naduhup & 1 & 0 & 0 & 0 \\
            Pidgin & 1 & 0 & 0 & 0 \\
            Baining & 1 & 0 & 0 & 0 \\
            Blue Nile Mao & 1 & 0 & 0 & 0 \\
            Furan & 1 & 0 & 0 & 0 \\
            Nyimang & 1 & 0 & 0 & 0 \\
            Marrku-Wurrugu & 1 & 0 & 0 & 0 \\
            Uru-Chipaya & 1 & 0 & 0 & 0 \\
            Huavean & 1 & 0 & 0 & 0 \\
            Mairasic & 1 & 0 & 0 & 0 \\
            Araucanian & 1 & 1 & 0 & 0 \\
            Shastan & 1 & 0 & 0 & 0 \\
            North Halmahera & 1 & 0 & 0 & 0 \\
            Tor-Orya & 1 & 0 & 0 & 0 \\
            Chapacuran & 1 & 0 & 0 & 0 \\
            Yuat & 1 & 0 & 0 & 0 \\
            Taulil-Butam & 1 & 0 & 0 & 0 \\
            Jicaquean & 1 & 0 & 0 & 0 \\
            \bottomrule
        \end{tabular}%
    }
    \caption{Distribution of languages in Glottolog language families across all tasks.}
    \label{tab:distribution_of_language_families_across_tasks_table3}
\end{table}

For all tasks, Indo-European languages are well-represented. Nuclear Trans New Guinea languages are well-represented in \WordTranslationAbbreviation, while Atlantic-Congo languages are well-represented in \WordTranslationAbbreviation~ and \WordTranslationInContextAbbreviation.

\subsection{Language code conventions}

In \WordTranslationAbbreviation, we represent each translation by its ISO code, regardless of the translation's script or geographic origin. For example, Achinese may be written in Arabic or Latin script. However, this distinction in script is not made in \WordTranslationAbbreviation~ as PanLex \textemdash a major data source\textemdash classifies word translations only by their ISO code. Consequently, translations to and from the language Achinese falls under the ISO code \ttt{ace}.

The data sources of \WordTranslationInContextAbbreviation, \FillInTheBlankAbbreviation, and \BagOfWordsMachineTranslationAbbreviation~ differentiate languages by script. For example, Achinese in Arabic script is evaluated separately from Achinese in Latin script. We adopt this distinction by script for these three tasks. 

\newpage

\section{Licensing} \label{sec:licensing}
The datasets used for this study are all publicly available. \textbf{FLORES+} is released under the Creative Commons Attribution-ShareAlike 4.0 International Public License. \textbf{GLOTLID} is released under Apache 2.0. \textbf{GATITOS} is released under Creative Commons Attribution 4.0. \textbf{IDS} is released under the Creative Commons license. While \textbf{PanLex} is licensed under Creative Commons CC0 1.0 Universal, PanLex draws upon numerous sources, each of which has its own copyright status. Under Creative Commons CC0 1.0 Universal, the use of PanLex for research purposes is permitted.

\section{Lexicons} \label{sec:lexicons}
In \autoref{tab:lexicon_source_breakdown}, we detail the number of languages covered in our three lexicon sources. PanLex covers the most.

\begin{table}[!hbtp]
    \centering
    \small
    \begin{tabular}{c|c}
        \toprule
         \textbf{Lexicon Source} & \textbf{\# of Language Pairs} \\
         \midrule
         GATITOS & 177 \\
         IDS & 240 \\
         PanLex & 5731 \\
         \bottomrule
    \end{tabular}
    \caption{Breakdown of languages across lexicon sources}
    \label{tab:lexicon_source_breakdown}
\end{table}

\section{Classification Heuristics} \label{sec:classification_heuristics}
\subsection{Examples across tasks}
See \autoref{tab:word_translation_examples}, \autoref{tab:word_translation_in_context_examples}, \autoref{tab:fill_in_the_blank_examples}, and \autoref{tab:bow_mt_examples} for examples of responses that were classified as correct and incorrect. We further define the implementation of each classification heuristic.

\begin{table*}[hbtp]
    \centering
    \small
    \begin{tabular}{P{7cm} | P{2.5cm} | P{2cm} | P{2.5cm} }
        \toprule
        \textbf{Prompt} & \textbf{Model Response} & \textbf{Ground Truth} & \textbf{Classification} \\
        \midrule
        \multicolumn{4}{l}{\textbf{Correct}} \\
        \midrule
        \makecell[tl]{Translate the following word from Papiamento to \\English. Respond with a single word.\\ Word:conoci. \\Translation:} & know & know & exact match \\ \midrule
        Translate the following text from English to Spanish: Egyptian. & Egipto. & egipcio, de egipto & inflection \\ \midrule
        Translate the following text from English to Spanish: Charm. & El encanto. & encanto & substring \\ \midrule
        \makecell[tl]{Translate the following word from French to English.\\Respond with a single word.\\Word:un des quadruplés.\\Translation:} & one of the quadruplets & quad, quadruplet & inflection in substring \\ \midrule
        \makecell[tl]{Translate the following word from Dutch to English.\\Respond with a single word.\\Word:smerig.\\Translation:} & disgusting & dirty, soiled, squalid & synonym \\ \midrule
        \multicolumn{4}{l}{\textbf{Incorrect}} \\
        \midrule
        Translate the following text from English to Yagaria: hill. & hill & ae, moa, aemo & echo \\ \midrule
        \makecell[tl]{Translate the following word from English to Gane.\\Respond with a single word.\\Word: road\\Translation:}& path & lolan & outputted in source language \\ \midrule
        Translate the following text from Djamindjung to English: mayili. &  mayil. mayil. mayil. mayil. mayil. & poison & gibberish \\
        \bottomrule
    \end{tabular}
    \caption{Examples of \WordTranslationAbbreviation~ prompts and responses.}
    \label{tab:word_translation_examples}
\end{table*}

\begin{table*}[hbtp]
    \centering
    \small
    \begin{tabular}{P{7.5cm} | P{2.5cm} | P{2cm} | P{2cm} }
        \toprule
        \textbf{Prompt} & \textbf{Model Response} & \textbf{Ground Truth} & \textbf{Classification} \\
        \midrule
        \multicolumn{4}{l}{\textbf{Correct}} \\
        \midrule
        In ``Zot dir li ankor, ``Kisannla to ete? Rann enn repons pou nou al inform bann dimoun ki finn avoy nou. Twa, ki to ena pou dir lor tomem?'''', the word ``ankor'' means \_\_\_\_ in English. & again & again, still & exact match \\ \midrule
        In ``A Comissão tem de dar apoio a esses países mas deve também utilizá-los como exemplo para obrigar outros países a fazerem o mesmo.'', the word ``países'' means \_\_\_\_ in English. & countries & country & inflection \\ \midrule
        In ``Around 11:29, the protest moved up Whitehall, past Trafalgar Square, along the Strand, passing by Aldwych and up Kingsway towards Holborn where the Conservative Party were holding their Spring Forum in the Grand Connaught Rooms hotel.'', the word ``were'' means \_\_\_\_ in Hungarian. & a) voltak & voltak & substring \\ \midrule
        \makecell[tl]{Sentence: Dit moedigt men aan om zelfs veur aafstande\\vaan `n paar honderd meter de auto te nómme (aafstande\\die door `t wegenetwerk `n paar kilometer kenne wère).\\ \\Define ``auto'' in one English word: } & cars, trucks, motorcycles, bicycles, scooters, mopeds, motorbikes & car & inflection in substring \\ \midrule
        ``In ``Laylak ez zituen arrosak espero.'', the word ``zituen'' means \_\_\_\_ in English.'' & they will have & had & synonym \\ \midrule
        \multicolumn{4}{l}{\textbf{Incorrect}} \\ \midrule
        What does ``in'' mean in Kambera in the sentence ``She didn’t trust in the LORD.''? Meaning (one word): & in & hudalu, coda, hu dalu, nu dalu & echo \\ \midrule
        In ``The protest started around 11:00 local time (UTC+1) on Whitehall opposite the police-guarded entrance to Downing Street, the Prime Minister's official residence.'', the word ``time'' means \_\_\_\_ in Hmong Daw. & The protest started around 11:00 local time (UTC+1) on Whitehall opposite the police-guarded entrance to Downing Street, the Prime Minister's official residence. & sij hawm & outputted source sentence \\ \midrule
        In ``The couple has to fill an application form and submit it along with two photographs of their wedding ceremony and an invitation card for the same.'', the word ``same'' means \_\_\_\_ in Akawaio. & similar & mari & outputted in source language \\ \midrule
        In ``Christopher Garcia, a spokesperson of the Los Angeles Police Department, said the suspected male offender is being investigated for trespassing rather than vandalism.'', the word ``male'' means \_\_\_\_ in Mountain Koiali. & gender & mo, ovaite & gibberish \\ \bottomrule
    \end{tabular}
    \caption{Examples of \WordTranslationInContextAbbreviation~ prompts and response}
    \label{tab:word_translation_in_context_examples}
\end{table*}

\begin{table*}[hbtp]
    \centering
    \small
    \begin{tabular}{P{10cm} | P{2.5cm} | P{2cm} }
        \toprule
        \textbf{Prompt} & \textbf{Next Word} & \textbf{Probability} \\
        \midrule
        \makecell[tl]{Translate the sentence into English.\\Standard Latvian:Mums tagad ir 4 mēnešus vecas peles, kas nav diabēta\\slimnieces, bet kuras agrāk bija diabēta slimnieces, viņš piebilda.\\English: ``We now have 4-month-old} & mice & $2.37 \times 10^{-4}$ \\
        \midrule
        \makecell[tl]{Translate the sentence into English. \\Czech:Dr. Ehud Ur, profesor medicíny na Dalhousieově univerzitě v\\ Halifaxu v Novém Skotsku a zároveň předseda klinické a vědecké divize\\ Kanadské diabetické asociace upozornil, že výzkum je teprve ve svých \\počátcích. \\ English: Dr. Ehud Ur,} & professor & 0.367 \\
        \midrule
        \makecell[tl]{Translate the following text into Sundanese.\\ English:USA Gymnastics and the USOC have the same goal — making the\\sport of gymnastics, and others, as safe as possible for athletes to follow their\\dreams in a safe, positive and empowered environment. \\ Sundanese: Organisasi Senam Nasional AS sareng USOC gaduh tujuan anu\\sami - ngusahakeun olahraga} & sename & 0.903 \\
        \bottomrule
    \end{tabular}
    \caption{Examples of \FillInTheBlankAbbreviation~ prompts. LLMs may be surprisingly brittle to punctuation and spacing, but we find that these result in minor variations of our results.}
    \label{tab:fill_in_the_blank_examples}
\end{table*}

\begin{table*}[hbtp]
    \centering
    \small
    \begin{tabular}{P{5cm} | P{5cm} | P{5cm} }
        \toprule
        \textbf{Prompt} & \textbf{Model Response} & \textbf{Ground Truth} \\
        \midrule
        Translate into English: Edgar Vagwilla wa li valula ko kuoko kuenda ko njembue, osimbu okuti Kristoffer Schneider wa sukila oku pelaliwa vocipala. & \textcolor{darkgreen}{Edgar} \textcolor{violet}{Vagwilla} \textcolor{darkgreen}{was} \textcolor{blue}{injured} in the knee to the ankle, \textcolor{darkgreen}{while Kristoffer Schneider} \textcolor{darkgreen}{was} \textcolor{blue}{taken} to the hospital \textcolor{darkgreen}{for} a knee injury. & Edgar Veguilla received arm and jaw wounds while Kristoffer Schneider was left requiring reconstructive surgery for his face. \\
        \midrule
        Translate into Kikuyu: Perry stated that he would ``return to Texas to assess the results of tonight's caucus, determine whether there is a path forward for myself in this race'', but later said that he would remain in the race and compete in the January 21 South Carolina primary. & \textcolor{darkgreen}{Perry} alitangaza kwamba angerudi \textcolor{darkgreen}{Texas} ``kuchunguza matokeo ya usiku huu \textcolor{darkgreen}{wa} kampeni, kuamua \textcolor{darkgreen}{kama} kuna njia ya \textcolor{violet}{mbele} kwangu katika mbio hii'', lakini baadaye alisema kwamba angebaki katika mbio \textcolor{darkgreen}{na} kushiriki katika January 21 \textcolor{darkgreen}{South Carolina} primary. & Perry akiuga atĩ we ``Nĩ egũcoka Texas kũrora maumĩrĩra ma mũcemanio wa atangoria ũtukũ ũcio, kũrora kana kwĩna gacĩra gakwa ga kũhotithia gũthiĩ na mbere gĩcindano-inĩ kĩu'', no thutha ũcio akiuga nĩ egũthiĩ na mbere na gĩcindano kĩu na nĩ egũcindana ithurano-inĩ cia kĩambĩrĩria mweri mĩrongo ĩrĩ na ũmwe South Carolina. \\
        \bottomrule
    \end{tabular}
    \caption{Examples of \BagOfWordsMachineTranslationAbbreviation~ prompts and responses. Exact matches on words are colored in \textcolor{darkgreen}{green}, inflections in \textcolor{violet}{violet}, and synonyms in \textcolor{blue}{blue}.}
    \label{tab:bow_mt_examples}
\end{table*}

\subsubsection{What counts as ``correct''} \label{subsubsec:correct_classification_heuristics}

\paragraph{\ttt{exact\_match}} If the model prediction matched any uncased, unpunctuated ground-truth answer, the prediction was marked as an \ttt{exact\_match}.

\paragraph{\ttt{inflection}} We make use of the Python package \ttt{fuzzywuzzy}, a package that uses Levenshtein distance to perform fuzzy string matching. We classify a model prediction as an \ttt{inflection} should it achieve a \ttt{fuzzywuzzy} \footnote{https://pypi.org/project/fuzzywuzzy/} similarity score of at least 75. %

\paragraph{\ttt{substring}} We mark a prediction as \ttt{substring} should any of the ground-truth answers exist as a word/phrase of the model's prediction, irrespective of punctuation or case.

\paragraph{\ttt{inflection\_within\_substring}} We denote a model prediction as \ttt{inflection\_within\_substring} if any inflected form of the ground truth, as defined above, is contained within the model prediction, ignoring punctuation and case.

\paragraph{\ttt{synonym}} We designate a model prediction as a \ttt{synonym} if it belongs to any \ttt{WordNet} synset of the ground truth answers. The usage of \ttt{WordNet} restricts this classification type to the \comprehension~ direction in \WordTranslationAbbreviation, \WordTranslationInContextAbbreviation, and \BagOfWordsMachineTranslationAbbreviation. 

\subsubsection{Error classification}
We designate the following categories of incorrect responses.
\paragraph{\ttt{echo}} A prediction is an \ttt{echo} if it matches the word to be translated, ignoring casing and punctuation.

\paragraph{\ttt{outputted\_in\_source\_language}} If the prediction does not satisfy any of the above classification types but can be found on the source side of a translation lexicon, the prediction is marked as \ttt{outputted\_in\_source\_language}.

\paragraph{\ttt{gibberish}} Should the prediction fail to fall into any of these classification type, the prediction is marked as \ttt{gibberish}.

\subsection{Manual Evaluation} \label{subsec:manual_evaluation}
\begin{table}[!hbtp]
    \centering
    \begin{tabular}{c|c|c}
         \toprule
         \textbf{Task} & \textbf{False Positive} & \textbf{False Negative}\\
         \midrule
         \WordTranslationAbbreviation & 2.5\% & 2.5\% \\
         \WordTranslationInContextAbbreviation & 10.7\% & 1.7\% \\
         \BagOfWordsMachineTranslationAbbreviation & 5.7\% & 1.7\% \\
         \bottomrule
    \end{tabular}
    \caption{To achieve these results, we evaluated 283 samples from \WordTranslationAbbreviation, 121 from \WordTranslationInContextAbbreviation, and 229 from \BagOfWordsMachineTranslationAbbreviation.}
    \label{tab:manual_evaluation}
\end{table}

To perform manual evaluation, we randomly selected a language for each model-evaluation direction pair and annotated at least 10 responses from it. \autoref{tab:manual_evaluation} highlights low false positives and negatives are across our evaluations. This suggests that the evaluation metrics applied to our models are reliable. %

\newpage

\section{Prompt Exploration} \label{sec:prompt_exploration}
We recognize that LLMs are sensitive to the prompts used for each task \citep{anagnostidis2024susceptiblellmsinfluenceprompts}.

We evaluated our six models on a series of ``candidate'' prompts: prompts that clearly delineate the word to translate as well as any additional context. We ran these small evaluations in Spanish as we assumed that if the model could not accurately perform the task in an HRL, such as Spanish, a model would be unlikely to do so in an LRL.

We list models and the candidate prompts they were matched with in the sections below. 
All our experiments use deterministic generation for decoding.

\subsection{Word Translation}
Our candidate prompts stress succinctness in the translation. We emphasized that the model translation be one word to make parsing simpler. \newline

\noindent\textbf{\comprehension}

\textbf{Prompt 1:} We assigned the prompt below to \ayaTwo, \falcon, and \llama~ for \WordTranslationAbbreviation~ in the \comprehension~ direction.
\begin{quote}\raggedright
    \ttt{Translate the following word from \{target language\} to English. Respond with a single word.} \newline
    \ttt{Word: \{word\}} \newline
    \ttt{Translation:}
\end{quote}

\textbf{Prompt 2:} We assigned the prompt below to \ayaOne~ and \bloom.
\begin{quote}\raggedright
    \ttt{Translate the following text from \{target language\} to English: \{word\}.}
\end{quote}

\textbf{Prompt 3:} We assigned the prompt below to \gemma.
\begin{quote}\raggedright
    \ttt{Translate `\{word\}' from \{target language\} into English. Respond in one word.}
\end{quote}

\noindent\textbf{\generation}

\textbf{Prompt 1:} We assigned the prompt below to \ayaTwo, \falcon, \llama.

\begin{quote}\raggedright
    \ttt{Translate the following word from English to \{target language\}. Respond with a single word.} \newline
    \ttt{Word: \{word\}} \newline
    \ttt{Translation:}
\end{quote}

\textbf{Prompt 2:} We assigned the prompt below to \ayaOne~ and \bloom.

\begin{quote}\raggedright
    \ttt{Translate the following text from English to \{target language\}: \{word\}.}
\end{quote}

\textbf{Prompt 3:} We assigned the prompt below to \gemma.

\begin{quote}\raggedright
    \ttt{Translate `\{word\}' from English to \{target language\}. Answer in one word:}
\end{quote}

\subsection{Word Translation with Context}
A common error we encountered involved models translating the entire sentence rather than a specific word. Consequently, our prompts emphasized translating a sole word. \newline

\noindent\textbf{\comprehension}

\textbf{Prompt 1:} We assign the prompt below to \ayaOne.

\begin{quote}\raggedright
    \ttt{What does `\{word\}' mean in English in the sentence `\{sentence\}'? Meaning (one word):}
\end{quote}

\textbf{Prompt 2}: We assign the prompt below to \ayaTwo~ and \falcon.

\begin{quote}\raggedright
    \ttt{In '\{sentence\}', the word `\{word\}' means \_\_\_\_ in English.}
\end{quote}

\textbf{Prompt 3}: We assign the prompt below to \bloom~ and \llama.

\begin{quote}\raggedright
    \ttt{Sentence: \{sentence\}} \newline
    \ttt{Define `\{word\}' in one English word: }
\end{quote}

\textbf{Prompt 4}: We assign the prompt to \gemma.

\begin{quote}\raggedright
    \ttt{Sentence: \{sentence\}} \newline
    \ttt{English definition of `\{word\}'}
\end{quote}

\noindent\textbf{\generation}

\textbf{Prompt 1}: We assign the prompt below to \ayaOne.

\begin{quote}\raggedright
    \ttt{What does `\{word\}' mean in \{target language\} in the sentence `\{sentence\}'? Meaning (one word): }
\end{quote}

\textbf{Prompt 2}: We assign the prompt to \ayaTwo, \falcon, \gemma, \llama.

\begin{quote}\raggedright
    \ttt{In `\{sentence\}', the word `\{word\}' means \_\_\_\_ in \{target language\}.}
\end{quote}

\textbf{Prompt 3}: We assign the prompt below to \bloom.

\begin{quote}\raggedright
    \ttt{Define `\{word\}' in `\{sentence\}' in \{target language\}:}
\end{quote}

\subsection{Translation-Conditioned Language Modeling}
Prompt construction depended on model architecture. Because \ayaOne~ uses an encoder-decoder architecture, the first $n$ words in the target translation are fed into the decoder rather than encoded as a prompt. The remaining five models utilized decoder architecture; the target translation of the first $n$ words was part of the prompt. \newline

\noindent\textbf{\comprehension}

\textbf{Prompt 1}: We assign the prompt below to \ayaOne.

\begin{quote}\raggedright
    \ttt{Translate the sentence into English:} \newline
    \ttt{\{Target Language\}:\{source sentence\}} \newline
    \ttt{English:}
\end{quote}

\textbf{Prompt 2}: We assign the prompt below to \ayaTwo, \bloom, \falcon, \gemma, and \llama. \newline

\begin{quote}\raggedright
    \ttt{Translate the sentence into English.} \newline
    \ttt{\{Target Language\}:\{source sentence\}} \newline
    \ttt{English: \{target translation up to index $n$\}}
\end{quote}

\noindent\textbf{\generation}

\textbf{Prompt 1}: We assign the prompt below to \ayaOne.

\begin{quote} \raggedright
    \ttt{Translate the following text into \{target language\}.} \newline
    \ttt{English: \{source sentence\}} \newline
    \ttt{\{Target Language\}}:
\end{quote}

\textbf{Prompt 2}: We assign the prompt below to \ayaTwo, \bloom, \gemma, \falcon, and \llama. 

\begin{quote} \raggedright
    \ttt{Translate the following text into \{target language\}.} \newline
    \ttt{English:\{source sentence\}} \newline
    \ttt{Target Language: \{target translation up to index $n$\}}
\end{quote}

\subsection{Bag-of-Words Machine Translation}
When prompted to translate a sentence, model outputs often missed the objective; models provided additional context to the subject of the sentence. To avoid confusion of what was expected, we made the act of translation as explicit as possible. \newline

\noindent\textbf{\comprehension}

\textbf{Prompt 1}: We assigned the prompt below to \gemma.

\begin{quote} \raggedright
    \ttt{Sentence: \{source sentence\}} \newline
    \ttt{English translation:}
\end{quote}

\textbf{Prompt 2}: We assigned the prompt below to \llama.

\begin{quote} \raggedright
    \ttt{What does this sentence mean in English: \{source sentence\}?}
\end{quote}

\textbf{Prompt 3}: We assigned the prompt below to \ayaOne, \ayaTwo, \bloom, and \falcon.

\begin{quote} \raggedright
    \ttt{Translate into English: \{source sentence\}}
\end{quote}

\noindent\textbf{\generation}

\textbf{Prompt 1}: We assigned the prompt below to \gemma.

\begin{quote}
    \ttt{Sentence: \{source sentence\}} \newline
    \ttt{\{Target Language\} translation:}
\end{quote}

\textbf{Prompt 2}: We assigned the prompt below to \llama.

\begin{quote}
    \ttt{English sentence: \{source sentence\}} \newline
    \ttt{\{Target Language\} translation:}
\end{quote}

\textbf{Prompt 3}: We assigned the prompt below to \ayaOne, \ayaTwo, \bloom, and \falcon.

\begin{quote}
    \ttt{Translation into \{target language\}: \{source sentence\}}
\end{quote}

\newpage

\section{Results in Detail} \label{sec:results_in_detail}
\subsection{Feature Importance} \label{subsec:feature_importance}
\begin{figure*}[!hbtp]
    \centering
    \small
    \resizebox{\textwidth}{!}{%
    \begin{forest}
        for tree={
            draw, %
            align=center, %
            edge={->},
            s sep=2pt,
            l sep=10pt,
        }
        [{model supports language \\samples=1900\\value=16.2}
            [{resource level $\le$ 66445.5 \\samples=279\\value=34.4}, 
                edge label={node[midway,xshift=-15pt,yshift=2pt]{True}}
                [{from \bloom\\samples=68\\value=15.5},
                    [{is \generation\\samples=36\\values=11.0},
                        [{samples=17\\values=5.8}],
                        [{samples=19\\values=15.6}]
                    ],
                    [{$\in$ Atlantic-Congo family\\samples=32\\values=20.7},
                        [{samples=13\\values=15.7}],
                        [{samples=19\\values=24.1}]
                    ]
                ]
                [{resource level $\le$ 1670427.5\\samples=211\\value=40.5},
                    [{resource level $\le$ 496354\\samples=108\\values=35.5},
                        [{samples=49\\values=31.5}],
                        [{samples=59\\values=38.9}]
                    ],
                    [{from \falcon\\samples=103\\values=45.7},
                        [{samples=15\\values=34.0}],
                        [{samples=88\\values=47.7}]
                    ]
                ]
            ]
            [{$\in$ Indo-European family\\samples=1621\\value=13.0},
                edge label={node[midway,xshift=15pt,yshift=2pt]{False}}
                [{from \falcon\\samples=596\\value=20.4},
                    [{uses Latin script\\samples=111\\values=9.2},
                        [{samples=63\\values=14.7}],
                        [{samples=48\\values=2.0}]
                    ],
                    [{is \generation\\samples=485\\values=22.9},
                        [{samples=240\\values=18.2}],
                        [{samples=245\\values=27.5}]
                    ]
                ]
                [{resource level $\le$ 58344.5\\samples=1025\\value=8.7},
                    [{is \generation\\samples=647\\values=5.2},
                        [{samples=321\\values=3.1}],
                        [{samples=326\\values=7.2}]
                    ],
                    [{from \llama\\samples=378\\values=14.8},
                        [{samples=73\\values=25.9}],
                        [{samples=305\\values=12.1}]
                    ]
                ]
            ]
        ]
    \end{forest}
    }
    \caption{A decision tree trained on linguistic and task features as well as \textbf{Word Translation} language scores.} %
    \label{tree:wt}
\end{figure*}
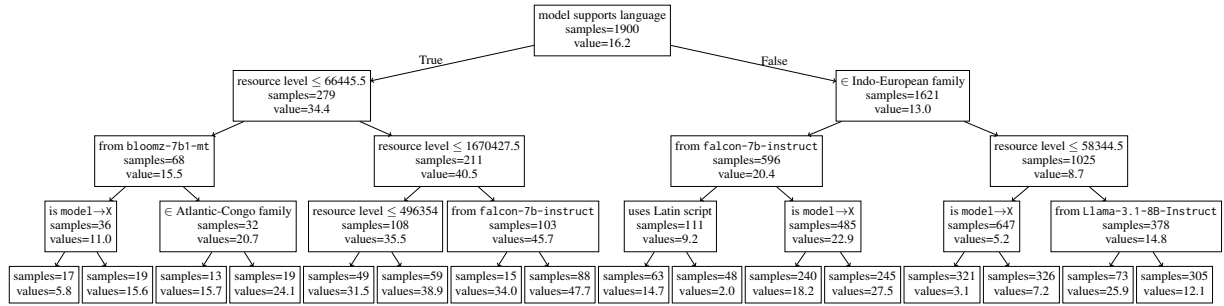

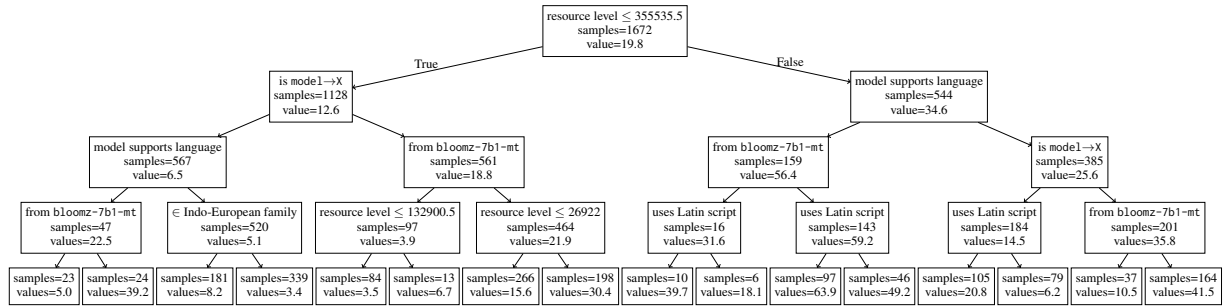
\begin{figure*}[hbtp]
    \centering
    \small
    \resizebox{\textwidth}{!}{%
    \begin{forest}
        for tree={
            draw, %
            align=center, %
            edge={->},
            s sep=2pt,
            l sep=10pt,
        }
        [{resource level $\le$ 355535.5\\samples=1672\\value=19.8}
            [{is $\generation$\\samples=1128\\value=12.6}, 
                edge label={node[midway,xshift=-15pt,yshift=2pt]{True}}
                [{model supports language\\samples=567\\value=6.5},
                    [{from \bloom\\samples=47\\values=22.5},
                        [{samples=23\\values=5.0}],
                        [{samples=24\\values=39.2}]
                    ],
                    [{$\in$ Indo-European family\\samples=520\\values=5.1},
                        [{samples=181\\values=8.2}],
                        [{samples=339\\values=3.4}]
                    ]
                ]
                [{from \bloom\\samples=561\\value=18.8},
                    [{resource level $\le$ 132900.5\\samples=97\\values=3.9},
                        [{samples=84\\values=3.5}],
                        [{samples=13\\values=6.7}]
                    ],
                    [{resource level $\le$ 26922 \\samples=464\\values=21.9},
                        [{samples=266\\values=15.6}],
                        [{samples=198\\values=30.4}]
                    ]
                ]
            ]
            [{model supports language\\samples=544\\value=34.6},
                edge label={node[midway,xshift=15pt,yshift=2pt]{False}}
                [{from \bloom\\samples=159\\value=56.4},
                    [{uses Latin script\\samples=16\\values=31.6},
                        [{samples=10\\values=39.7}],
                        [{samples=6\\values=18.1}]
                    ],
                    [{uses Latin script\\samples=143\\values=59.2},
                        [{samples=97\\values=63.9}],
                        [{samples=46\\values=49.2}]
                    ]
                ]
                [{is \generation\\samples=385\\value=25.6},
                    [{uses Latin script\\samples=184\\values=14.5},
                        [{samples=105\\values=20.8}],
                        [{samples=79\\values=6.2}]
                    ],
                    [{from \bloom\\samples=201\\values=35.8},
                        [{samples=37\\values=10.5}],
                        [{samples=164\\values=41.5}]
                    ]
                ]
            ]
        ]
    \end{forest}
    }
    \caption{A decision tree trained on linguistic and task features and \textbf{Word Translation with Context} language scores.} %
    \label{tree:wtic}
\end{figure*}

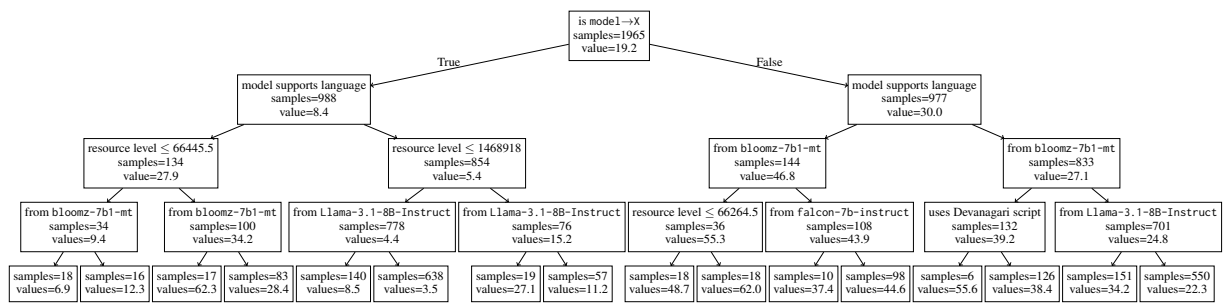
\begin{figure*}[hbtp]
    \centering
    \small
    \resizebox{\textwidth}{!}{%
    \begin{forest}
        for tree={
            draw, %
            align=center, %
            edge={->},
            s sep=2pt,
            l sep=10pt,
        }
        [{is \generation\\samples=1965\\value=19.2}
            [{model supports language\\samples=988\\value=8.4}, 
                edge label={node[midway,xshift=-15pt,yshift=2pt]{True}}
                [{resource level $\le$ 66445.5\\samples=134\\value=27.9},
                    [{from \bloom\\samples=34\\values=9.4},
                        [{samples=18\\values=6.9}],
                        [{samples=16\\values=12.3}]
                    ],
                    [{from \bloom\\samples=100\\values=34.2},
                        [{samples=17\\values=62.3}],
                        [{samples=83\\values=28.4}]
                    ]
                ]
                [{resource level $\le$ 1468918\\samples=854\\value=5.4},
                    [{from \llama\\samples=778\\values=4.4},
                        [{samples=140\\values=8.5}],
                        [{samples=638\\values=3.5}]
                    ],
                    [{from \llama\\samples=76\\values=15.2},
                        [{samples=19\\values=27.1}],
                        [{samples=57\\values=11.2}]
                    ]
                ]
            ]
            [{model supports language\\samples=977\\value=30.0},
                edge label={node[midway,xshift=15pt,yshift=2pt]{False}}
                [{from \bloom\\samples=144\\value=46.8},
                    [{resource level $\le$ 66264.5\\samples=36\\values=55.3},
                        [{samples=18\\values=48.7}],
                        [{samples=18\\values=62.0}]
                    ],
                    [{from \falcon\\samples=108\\values=43.9},
                        [{samples=10\\values=37.4}],
                        [{samples=98\\values=44.6}]
                    ]
                ]
                [{from \bloom\\samples=833\\value=27.1},
                    [{uses Devanagari script\\samples=132\\values=39.2},
                        [{samples=6\\values=55.6}],
                        [{samples=126\\values=38.4}]
                    ],
                    [{from \llama\\samples=701\\values=24.8},
                        [{samples=151\\values=34.2}],
                        [{samples=550\\values=22.3}]
                    ]
                ]
            ]
        ]
    \end{forest}
    }
    \caption{A decision tree trained on linguistic and task features as well as \textbf{Translation-Conditioned Language Modeling} language scores.} %
    \label{tree:fitb}
\end{figure*}

\begin{figure*}[hbtp]
    \centering
    \small
    \resizebox{\textwidth}{!}{%
    \begin{forest}
        for tree={
            draw, %
            align=center, %
            edge={->},
            s sep=2pt,
            l sep=10pt,
        }
        [{from \bloom\\samples=1965\\value=30.5}
            [{resource level $\le$ 3338198\\samples=331\\value=2.4}, 
                edge label={node[midway,xshift=-15pt,yshift=2pt]{True}}
                [{resource level $\le$ 290182.5\\samples=300\\value=1.5},
                    [{is \generation\\samples=224\\values=0.9},
                        [{samples=110\\values=1.5}],
                        [{samples=114\\values=0.3}]
                    ],
                    [{model supports language\\samples=76\\values=3.3},
                        [{samples=16\\values=7.3}],
                        [{samples=60\\values=2.2}]
                    ]
                ]
                [{model supports language\\samples=31\\value=10.5},
                    [{is \generation\\samples=11\\values=21.4},
                        [{samples=5\\values=37.5}],
                        [{samples=6\\values=8.1}]
                    ],
                    [{resource level $\le$ 6970180.5\\samples=20\\values=4.5},
                        [{samples=12\\values=2.5}],
                        [{samples=8\\values=7.4}]
                    ]
                ]
            ]
            [{is \generation\\samples=1634\\value=36.2},
                edge label={node[midway,xshift=15pt,yshift=2pt]{False}}
                [{model supports language\\samples=825\\value=23.1},
                    [{from \falcon\\samples=101\\values=49.4},
                        [{samples=10\\values=25.6}],
                        [{samples=91\\values=52.0}]
                    ],
                    [{uses Latin script\\samples=724\\values=19.5},
                        [{samples=442\\values=24.5}],
                        [{samples=282\\values=11.6}]
                    ]
                ]
                [{from \falcon\\samples=809\\value=49.5},
                    [{uses Latin script\\samples=160\\values=24.5},
                        [{samples=99\\values=36.0}],
                        [{samples=61\\values=5.7}]
                    ],
                    [{resource level $\le$ 129662\\samples=649\\values=55.7},
                        [{samples=257\\values=68.4}],
                        [{samples=392\\values=47.3}]
                    ]
                ]
            ]
        ]
    \end{forest}
    }
    \caption{A decision tree trained on linguistic and task features as well as \textbf{Bag-of-Words Machine Translation} language scores.}
    \label{tree:bow_mt}
\end{figure*}
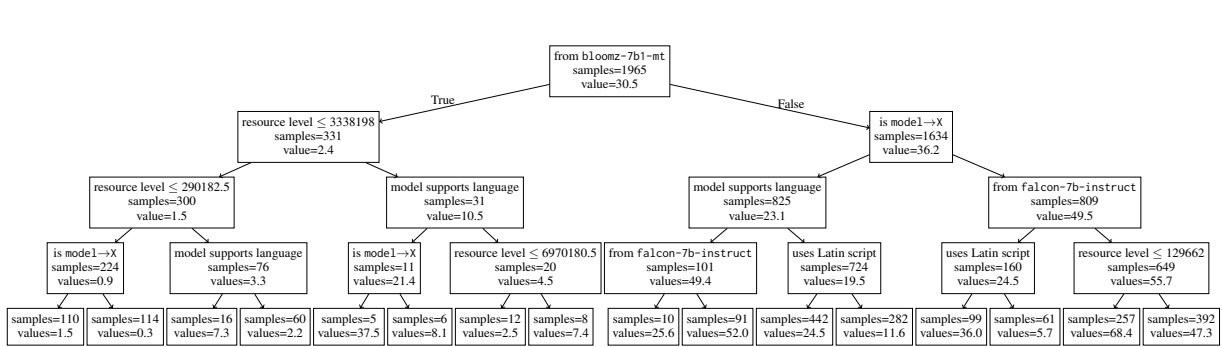

We trained a decision tree regressor on several features of a language: whether the model supports a language, the language's resource level (i.e. the number of Wikipedia pages available), which model predicted the language (e.g. \bloom, \llama, \falcon), which language family the language belonged to (e.g. Atlantic-Congo, Indo-European), what evaluation direction the model was assessed under, what script the language used (e.g. Latin), and the languages associated score. For task-specific decision trees, see \autoref{tree:wt}, \autoref{tree:wtic}, \autoref{tree:fitb}, and \autoref{tree:bow_mt}. \autoref{tab:avg_feature_importance_across_tasks} averages feature importance values and enumerates them in descending order.

\begin{table}[!hbtp]
    \centering
    \small
    \begin{tabular}{l|c}
        \toprule
        \textbf{Feature} & \makecell[c]{\textbf{Average Feature}\\\textbf{Importance}} \\
        \midrule
        Translation mode: \generation & $0.264 \pm 0.21$ \\ \midrule
        Supported by model & $0.2425 \pm 0.13$ \\ \midrule
        Resource level & $0.19 \pm 0.16$ \\ \midrule
        Model: \bloom & $0.148 \pm 0.14$ \\ \midrule
        Model: \falcon & $0.053 \pm 0.07$ \\ \midrule
        Family: Indo-European & $0.048 \pm 0.09$ \\ \midrule
        Script: Latin & $0.03 \pm 0.03$ \\ \midrule
        Model: \llama & $0.024 \pm 0.03$ \\ \midrule
        Script: Devanagari & $0.002 \pm 0.0$ \\
        \bottomrule
    \end{tabular}
    \caption{For each task, we trained a decision tree regressor with the language score as the label and attributes, such as the model in which the language was evaluated, the language's family, and its script, as features. Each regressor assigns importance scores for the features, ranging from 0 to 1 and reflecting their contribution to predicting the language's score. We then averaged feature importance across the four tasks and reported the features with non-zero importance scores. The overall average of each feature is depicted on the left of $\pm$ and the standard deviation on the right.}
    \label{tab:avg_feature_importance_across_tasks}
\end{table}

\newpage
\subsection{Model Averages}
\autoref{tab:average_model_performance_across_tasks} lists the model score averages across all tasks and evaluation directions.
\begin{table*}[!hbtp]
    \centering
    \small
    \resizebox{1\textwidth}{!}{
        \begin{tabular}{l | c c | c c | c c | c c }
            \toprule
            & \multicolumn{2}{c|}{\WordTranslationAbbreviation} & \multicolumn{2}{c|}{\WordTranslationInContextAbbreviation} & \multicolumn{2}{c|}{\FillInTheBlankAbbreviation} & \multicolumn{2}{c}{\BagOfWordsMachineTranslationAbbreviation} \\
             \midrule
            \textbf{Model} & \multicolumn{1}{c}{\textbf{\comprehension}} & \multicolumn{1}{c|}{\textbf{\generation}} 
            & \multicolumn{1}{c}{\textbf{\comprehension}} & \multicolumn{1}{c|}{\textbf{\generation}} 
            & \multicolumn{1}{c}{\textbf{\comprehension}} & \multicolumn{1}{c|}{\textbf{\generation}} 
            & \multicolumn{1}{c}{\textbf{\comprehension}} & \multicolumn{1}{c}{\textbf{\generation}} \\
            \midrule
            \ayaOne & 4.8 $\pm$ 10.9 & 3.4 $\pm$ 8.8 & \textbf{\underline{29.0 $\pm$ 22.3}} & \textbf{\underline{14.2 $\pm$ 19.8}} & 36.1 $\pm$ 12.2 & \textbf{\underline{14.2 $\pm$ 11.7}} & \textbf{\underline{65.6 $\pm$ 18.4}} & \textbf{\underline{35.9 $\pm$ 19.0}} \\
            \ayaTwo & 4.8 $\pm$ 10.2 & 2.8 $\pm$ 8.7 & 21.0 $\pm$ 22.2 & 7.2 $\pm$ 13.9 & 26.2 $\pm$ 12.3 & 7.2 $\pm$ 10.7 & 52.9 $\pm$ 21.2 & 26.4 $\pm$ 20.6 \\
            \bloom & 9.0 $\pm$ 8.2 & 2.1 $\pm$ 6.3 & 3.8 $\pm$ 6.0 & 3.6 $\pm$ 7.8 & \textbf{\underline{42.7 $\pm$ 12.1}} & 10.2 $\pm$ 18.4 & 1.1 $\pm$ 2.5 & 3.7 $\pm$ 8.5 \\
            \falcon & 2.8 $\pm$ 5.3 & 1.3 $\pm$ 4.8 & 12.2 $\pm$ 15.9 & 4.9 $\pm$ 9.4 & 17.6 $\pm$ 7.0 & 1.9 $\pm$ 5.0 & 25.5 $\pm$ 19.4 & 10.8 $\pm$ 8.5 \\
            \gemma & 4.1 $\pm$ 6.2 & 1.5 $\pm$ 4.6 & 18.0 $\pm$ 14.7 & 2.2 $\pm$ 5.0 & 22.9 $\pm$ 9.5 & 5.0 $\pm$ 6.5 & 48.0 $\pm$ 16.5 & 17.5 $\pm$ 13.4 \\
            \llama & \textbf{\underline{14.0 $\pm$ 8.7}} & \textbf{\underline{3.6 $\pm$ 7.9}} & 15.3 $\pm$ 15.9 & 6.7 $\pm$ 10.6 & 35.3 $\pm$ 12.5 & 11.7 $\pm$ 11.5 & 58.0 $\pm$ 15.7 & 26.4 $\pm$ 15.5 \\
            \bottomrule
        \end{tabular}
    }
    \caption{Average model accuracy percentages across tasks and comprehension (\texttt{X $\rightarrow$ eng}) and generative settings (\texttt{eng $\rightarrow$ X}) using the evaluation metrics defined in \autoref{sec:classification_heuristics}. The best-performing models for each task and direction are \textbf{bolded} and \underline{underlined}.}
    \label{tab:average_model_performance_across_tasks}
\end{table*}

\subsection{Language Family Averages}
\autoref{fig:performance_across_lang_families} shows for each task, the \emph{best} language family average across six models. We show language family averages across all tasks and models in \autoref{tab:lang_fam_wt}, \autoref{tab:lang_fam_wtic}, \autoref{tab:lang_fam_fitb}, and \autoref{tab:lang_fam_bow_mt}. While the Indo-European language family's average tends to be higher, there is more variation within the models themselves. In \WordTranslationAbbreviation~ \comprehension, \ayaOne's Turkic language family average is 11.7\% higher than \falcon's Indo-European language family average.

\begin{table*}[!hbtp]
    \centering
    \resizebox{1\textwidth}{!}{
    \begin{tabular}{l | c c | c c | c c | c c | c c}
        \toprule
        & \multicolumn{2}{c|}{\textbf{Indo-European}} & \multicolumn{2}{c|}{\textbf{Turkic}} & \multicolumn{2}{c|}{\textbf{Austronesian}} & \multicolumn{2}{c|}{\textbf{Afro-Asiatic}} & \multicolumn{2}{c}{\textbf{Atlantic-Congo}} \\
        \midrule
        \textbf{Model} & \multicolumn{1}{c}{\textbf{\comprehension}} & \multicolumn{1}{c|}{\textbf{\generation}} & \multicolumn{1}{c}{\textbf{\comprehension}} & \multicolumn{1}{c|}{\textbf{\generation}} & \multicolumn{1}{c}{\textbf{\comprehension}} & \multicolumn{1}{c|}{\textbf{\generation}} & \multicolumn{1}{c}{\textbf{\comprehension}} & \multicolumn{1}{c|}{\textbf{\generation}} & \multicolumn{1}{c}{\textbf{\comprehension}} & \multicolumn{1}{c}{\textbf{\generation}} \\ 
        \midrule
        \ayaOne & 27.4 $\pm$ 17.6 & 20.0 $\pm$ 16.9 & 23.4 $\pm$ 12.4 & 10.2 $\pm$ 13.5 & 3.6 $\pm$ 6.7 & 3.9 $\pm$ 7.6 & 5.7 $\pm$ 13.9 & 3.4 $\pm$ 9.8 & 2.0 $\pm$ 4.0 & 1.2 $\pm$ 3.0 \\
        \ayaTwo & 25.2 $\pm$ 17.8 & 20.0 $\pm$ 18.0 & 14.5 $\pm$ 9.2 & 4.8 $\pm$ 8.4 & 4.1 $\pm$ 6.6 & 2.7 $\pm$ 7.1 & 5.7 $\pm$ 14.5 & 3.2 $\pm$ 10.9 & 1.7 $\pm$ 2.4 & 0.4 $\pm$ 1.0 \\
        \bloom & 21.9 $\pm$ 16.7 & 11.3 $\pm$ 13.6 & 9.9 $\pm$ 4.2 & 1.3 $\pm$ 2.4 & 7.7 $\pm$ 5.5 & 3.1 $\pm$ 5.7 & 9.9 $\pm$ 10.6 & 2.4 $\pm$ 9.2 & 8.7 $\pm$ 4.1 & 0.7 $\pm$ 1.8 \\
        \falcon & 11.7 $\pm$ 11.7 & 9.3 $\pm$ 13.2 & 3.6 $\pm$ 2.2 & 1.6 $\pm$ 3.0 & 2.5 $\pm$ 2.8 & 0.8 $\pm$ 1.7 & 1.8 $\pm$ 1.6 & 0.3 $\pm$ 0.9 & 1.7 $\pm$ 1.9 & 0.3 $\pm$ 0.8 \\
        \gemma & 15.6 $\pm$ 13.1 & 10.0 $\pm$ 12.0 & 8.2 $\pm$ 4.6 & 1.6 $\pm$ 2.7 & 3.1 $\pm$ 3.1 & 1.0 $\pm$ 2.4 & 3.7 $\pm$ 4.1 & 0.6 $\pm$ 1.5 & 2.4 $\pm$ 2.5 & 0.5 $\pm$ 1.0 \\
        \llama & 29.3 $\pm$ 14.5 & 19.1 $\pm$ 14.8 & 22.1 $\pm$ 9.0 & 9.0 $\pm$ 9.8 & 13.9 $\pm$ 6.5 & 4.2 $\pm$ 6.7 & 15.2 $\pm$ 11.1 & 3.4 $\pm$ 8.4 & 11.2 $\pm$ 4.5 & 1.4 $\pm$ 2.1 \\
        \bottomrule
    \end{tabular}}
    \caption{Averaging scores by language family, model, and evaluation direction (i.e. \comprehension~ or \generation) for the task \textbf{Word Translation}. Data is written in the format \ttt{mean} $\pm$ \ttt{standard deviation}.}
    \label{tab:lang_fam_wt}
\end{table*}

\begin{table*}[!hbtp]
    \centering
    \resizebox{1\textwidth}{!}{
    \begin{tabular}{l | c c | c c | c c | c c | c c}
        \toprule
        & \multicolumn{2}{c|}{\textbf{Indo-European}} & \multicolumn{2}{c|}{\textbf{Turkic}} & \multicolumn{2}{c|}{\textbf{Austronesian}} & \multicolumn{2}{c|}{\textbf{Afro-Asiatic}} & \multicolumn{2}{c}{\textbf{Atlantic-Congo}} \\
        \midrule
        \textbf{Model} & \multicolumn{1}{c}{\textbf{\comprehension}} & \multicolumn{1}{c|}{\textbf{\generation}} & \multicolumn{1}{c}{\textbf{\comprehension}} & \multicolumn{1}{c|}{\textbf{\generation}} & \multicolumn{1}{c}{\textbf{\comprehension}} & \multicolumn{1}{c|}{\textbf{\generation}} & \multicolumn{1}{c}{\textbf{\comprehension}} & \multicolumn{1}{c|}{\textbf{\generation}} & \multicolumn{1}{c}{\textbf{\comprehension}} & \multicolumn{1}{c}{\textbf{\generation}} \\ 
        \midrule
        \ayaOne & 49.1 $\pm$ 17.4 & 30.4 $\pm$ 22.3 & 47.4 $\pm$ 15.5 & 12.0 $\pm$ 17.4 & 22.8 $\pm$ 18.2 & 10.1 $\pm$ 15.1 & 30.6 $\pm$ 23.2 & 21.9 $\pm$ 22.5 & 19.6 $\pm$ 12.9 & 5.7 $\pm$ 10.6 \\
        \ayaTwo & 43.8 $\pm$ 19.6 & 17.1 $\pm$ 18.9 & 24.8 $\pm$ 16.9 & 4.5 $\pm$ 10.8 & 15.2 $\pm$ 16.6 & 4.3 $\pm$ 10.3 & 24.5 $\pm$ 26.7 & 10.8 $\pm$ 18.1 & 6.2 $\pm$ 4.3 & 1.0 $\pm$ 1.3 \\
        \bloom & 8.3 $\pm$ 8.6 & 8.8 $\pm$ 12.4 & 1.2 $\pm$ 0.9 & 0.9 $\pm$ 1.5 & 2.4 $\pm$ 4.4 & 2.0 $\pm$ 4.1 & 3.4 $\pm$ 3.6 & 4.4 $\pm$ 9.3 & 2.3 $\pm$ 3.0 & 1.2 $\pm$ 2.5 \\
        \falcon & 27.2 $\pm$ 21.0 & 13.0 $\pm$ 14.6 & 4.9 $\pm$ 5.2 & 1.8 $\pm$ 2.8 & 9.6 $\pm$ 9.6 & 2.6 $\pm$ 3.9 & 6.6 $\pm$ 5.1 & 1.9 $\pm$ 3.1 & 4.7 $\pm$ 2.2 & 1.1 $\pm$ 1.0 \\
        \gemma & 31.8 $\pm$ 16.3 & 5.9 $\pm$ 8.1 & 14.8 $\pm$ 8.8 & 0.6 $\pm$ 1.3 & 13.7 $\pm$ 10.8 & 1.0 $\pm$ 2.5 & 19.4 $\pm$ 15.9 & 0.5 $\pm$ 1.0 & 9.9 $\pm$ 4.2 & 0.4 $\pm$ 0.5 \\
        \llama & 31.2 $\pm$ 16.3 & 14.8 $\pm$ 15.0 & 19.2 $\pm$ 14.1 & 3.2 $\pm$ 6.3 & 10.6 $\pm$ 11.3 & 5.0 $\pm$ 6.6 & 18.2 $\pm$ 16.0 & 2.7 $\pm$ 3.7 & 5.4 $\pm$ 4.5 & 2.4 $\pm$ 2.0 \\
        \bottomrule
    \end{tabular}}
    \caption{Averaging scores by language family, model, and evaluation direction (i.e. \comprehension~ or \generation) for the task \textbf{Word Translation with Context}. Data is written in the format \ttt{mean} $\pm$ \ttt{standard deviation}.}
    \label{tab:lang_fam_wtic}
\end{table*}

\begin{table*}[!hbtp]
    \centering
    \resizebox{1\textwidth}{!}{
    \begin{tabular}{l | c c | c c | c c | c c | c c}
        \toprule
        & \multicolumn{2}{c|}{\textbf{Indo-European}} & \multicolumn{2}{c|}{\textbf{Turkic}} & \multicolumn{2}{c|}{\textbf{Austronesian}} & \multicolumn{2}{c|}{\textbf{Afro-Asiatic}} & \multicolumn{2}{c}{\textbf{Atlantic-Congo}} \\
        \midrule
        \textbf{Language} & \multicolumn{1}{c}{\textbf{\comprehension}} & \multicolumn{1}{c|}{\textbf{\generation}} & \multicolumn{1}{c}{\textbf{\comprehension}} & \multicolumn{1}{c|}{\textbf{\generation}} & \multicolumn{1}{c}{\textbf{\comprehension}} & \multicolumn{1}{c|}{\textbf{\generation}} & \multicolumn{1}{c}{\textbf{\comprehension}} & \multicolumn{1}{c|}{\textbf{\generation}} & \multicolumn{1}{c}{\textbf{\comprehension}} & \multicolumn{1}{c}{\textbf{\generation}} \\ 
        \midrule
        \ayaOne & 43.7 $\pm$ 4.6 & 21.7 $\pm$ 11.6 & 37.3 $\pm$ 6.9 & 10.4 $\pm$ 9.0 & 35.9 $\pm$ 9.8 & 12.0 $\pm$ 8.2 & 36.6 $\pm$ 12.9 & 14.8 $\pm$ 10.0 & 24.8 $\pm$ 11.3 & 4.7 $\pm$ 4.8 \\
        \ayaTwo & 32.8 $\pm$ 11.4 & 12.0 $\pm$ 13.2 & 22.8 $\pm$ 6.7 & 3.0 $\pm$ 6.0 & 23.7 $\pm$ 8.9 & 5.7 $\pm$ 7.1 & 30.0 $\pm$ 15.9 & 13.1 $\pm$ 13.9 & 14.7 $\pm$ 1.2 & 1.2 $\pm$ 0.4 \\
        \bloom & 46.8 $\pm$ 11.9 & 16.2 $\pm$ 23.7 & 33.0 $\pm$ 1.7 & 0.5 $\pm$ 0.2 & 39.3 $\pm$ 7.8 & 6.4 $\pm$ 11.0 & 43.3 $\pm$ 16.1 & 9.3 $\pm$ 10.6 & 44.4 $\pm$ 8.7 & 5.4 $\pm$ 5.1 \\
        \falcon & 20.5 $\pm$ 9.8 & 3.9 $\pm$ 8.0 & 15.3 $\pm$ 0.6 & 0.2 $\pm$ 0.1 & 17.3 $\pm$ 3.8 & 1.9 $\pm$ 1.6 & 14.4 $\pm$ 1.8 & 0.4 $\pm$ 0.2 & 14.2 $\pm$ 1.2 & 0.7 $\pm$ 0.2 \\
        \gemma & 27.5 $\pm$ 9.9 & 8.1 $\pm$ 8.5 & 19.7 $\pm$ 3.8 & 1.5 $\pm$ 2.1 & 20.6 $\pm$ 7.2 & 4.8 $\pm$ 4.8 & 23.3 $\pm$ 9.0 & 4.3 $\pm$ 3.6 & 14.8 $\pm$ 1.5 & 1.5 $\pm$ 0.6 \\
        \llama & 43.9 $\pm$ 6.3 & 19.2 $\pm$ 12.1 & 35.7 $\pm$ 5.9 & 7.8 $\pm$ 5.6 & 32.8 $\pm$ 11.0 & 8.4 $\pm$ 7.7 & 34.6 $\pm$ 14.0 & 11.3 $\pm$ 9.9 & 20.3 $\pm$ 5.3 & 2.4 $\pm$ 2.1 \\
        
        \bottomrule
    \end{tabular}}
    \caption{Averaging scores by language family, model, and evaluation direction (i.e. \comprehension~ or \generation) for the task \textbf{Translation-Conditioned Language Modeling}. Data is written in the format \ttt{mean} $\pm$ \ttt{standard deviation}.}
    \label{tab:lang_fam_fitb}
\end{table*}

\begin{table*}[!hbtp]
    \centering
    \resizebox{1\textwidth}{!}{
    \begin{tabular}{l | c c | c c | c c | c c | c c}
        \toprule
         & \multicolumn{2}{c|}{\textbf{Indo-European}} & \multicolumn{2}{c|}{\textbf{Turkic}} & \multicolumn{2}{c|}{\textbf{Austronesian}} & \multicolumn{2}{c|}{\textbf{Afro-Asiatic}} & \multicolumn{2}{c}{\textbf{Atlantic-Congo}} \\
        \midrule
        \textbf{Language} & \multicolumn{1}{c}{\textbf{\comprehension}} & \multicolumn{1}{c|}{\textbf{\generation}} & \multicolumn{1}{c}{\textbf{\comprehension}} & \multicolumn{1}{c|}{\textbf{\generation}} & \multicolumn{1}{c}{\textbf{\comprehension}} & \multicolumn{1}{c|}{\textbf{\generation}} & \multicolumn{1}{c}{\textbf{\comprehension}} & \multicolumn{1}{c|}{\textbf{\generation}} & \multicolumn{1}{c}{\textbf{\comprehension}} & \multicolumn{1}{c}{\textbf{\generation}} \\ 
        \midrule
        \ayaOne & 75.3 $\pm$ 10.0 & 45.4 $\pm$ 16.4 & 69.0 $\pm$ 10.1 & 25.0 $\pm$ 20.4 & 57.7 $\pm$ 24.5 & 39.3 $\pm$ 22.6 & 60.8 $\pm$ 18.2 & 35.7 $\pm$ 18.4 & 56.2 $\pm$ 12.2 & 26.5 $\pm$ 10.5 \\
        \ayaTwo & 65.1 $\pm$ 17.9 & 36.2 $\pm$ 21.9 & 48.8 $\pm$ 11.6 & 16.0 $\pm$ 14.8 & 49.9 $\pm$ 19.1 & 31.5 $\pm$ 21.7 & 53.5 $\pm$ 26.1 & 33.3 $\pm$ 25.7 & 33.1 $\pm$ 4.6 & 15.9 $\pm$ 5.4 \\
        \bloom & 2.1 $\pm$ 3.6 & 5.2 $\pm$ 9.3 & 0.0 $\pm$ 0.0 & 0.5 $\pm$ 0.6 & 0.6 $\pm$ 1.9 & 5.5 $\pm$ 12.8 & 1.1 $\pm$ 1.4 & 0.5 $\pm$ 0.5 & 0.3 $\pm$ 0.4 & 1.6 $\pm$ 1.4 \\
        \falcon & 30.2 $\pm$ 25.2 & 12.8 $\pm$ 11.2 & 14.8 $\pm$ 10.6 & 4.3 $\pm$ 3.4 & 31.8 $\pm$ 16.0 & 14.6 $\pm$ 6.7 & 12.5 $\pm$ 10.1 & 4.9 $\pm$ 4.7 & 26.2 $\pm$ 4.3 & 11.6 $\pm$ 3.5 \\
        \gemma & 56.9 $\pm$ 16.5 & 22.8 $\pm$ 16.1 & 42.4 $\pm$ 7.8 & 8.4 $\pm$ 7.7 & 45.1 $\pm$ 14.8 & 23.7 $\pm$ 14.3 & 44.0 $\pm$ 16.8 & 11.3 $\pm$ 5.8 & 34.8 $\pm$ 4.2 & 14.4 $\pm$ 4.1 \\
        \llama & 67.8 $\pm$ 9.8 & 34.2 $\pm$ 15.1 & 62.2 $\pm$ 6.5 & 17.2 $\pm$ 10.1 & 57.3 $\pm$ 15.1 & 33.0 $\pm$ 16.8 & 53.5 $\pm$ 20.1 & 20.4 $\pm$ 10.9 & 45.1 $\pm$ 5.9 & 20.1 $\pm$ 5.7 \\
        \bottomrule
    \end{tabular}}
    \caption{Averaging scores by language family, model, and evaluation (i.e. \comprehension~ or \generation) for the task \textbf{Bag-of-Words Machine Translation}. Data is written in the format \ttt{mean} $\pm$ \ttt{standard deviation}.}
    \label{tab:lang_fam_bow_mt}
\end{table*}

\subsection{Resourceness} \label{subsec:resourceness}

\begin{figure*}[htbp]
    \centering
    \begin{minipage}{0.48\textwidth}
        \centering
        \includegraphics[width=\linewidth]{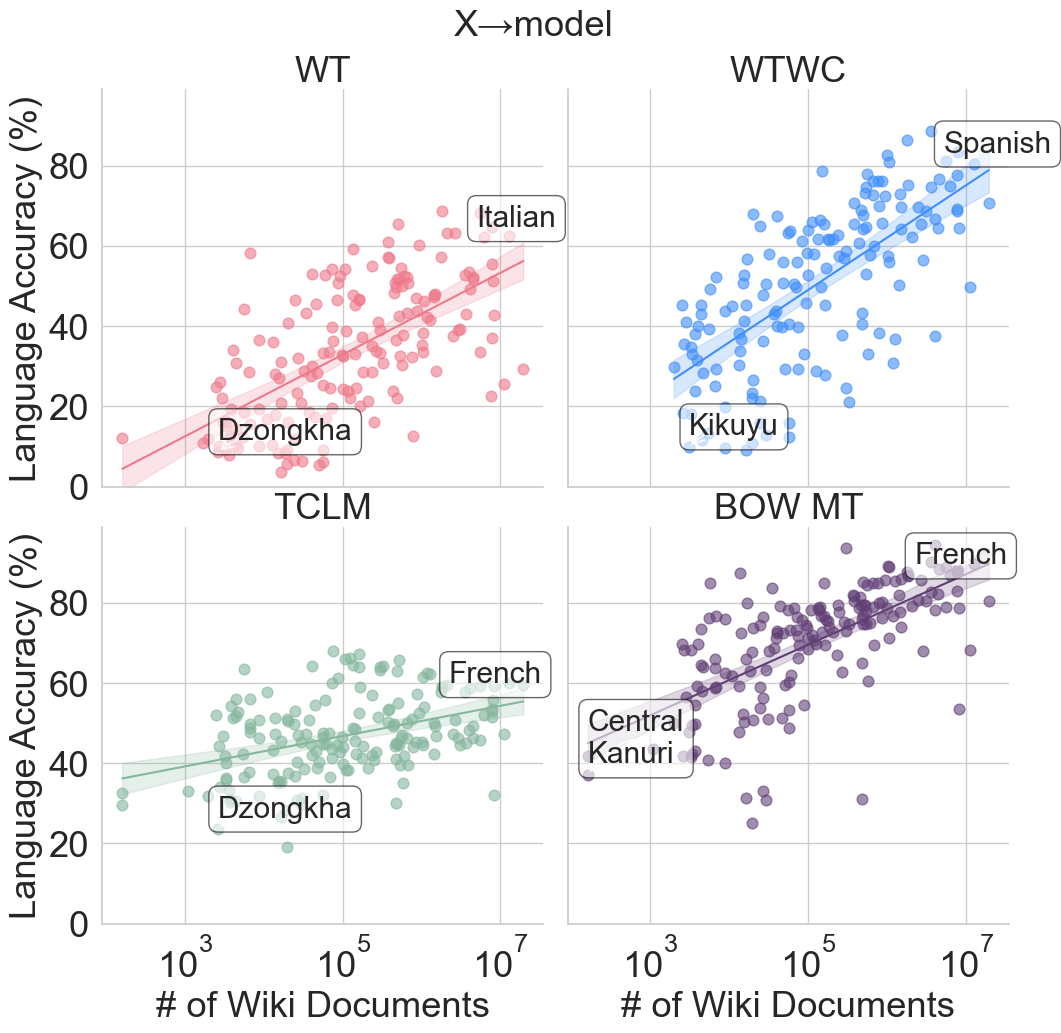}
    \end{minipage}%
    \hfill
    \begin{minipage}{0.48\textwidth}
        \centering
        \includegraphics[width=\linewidth]{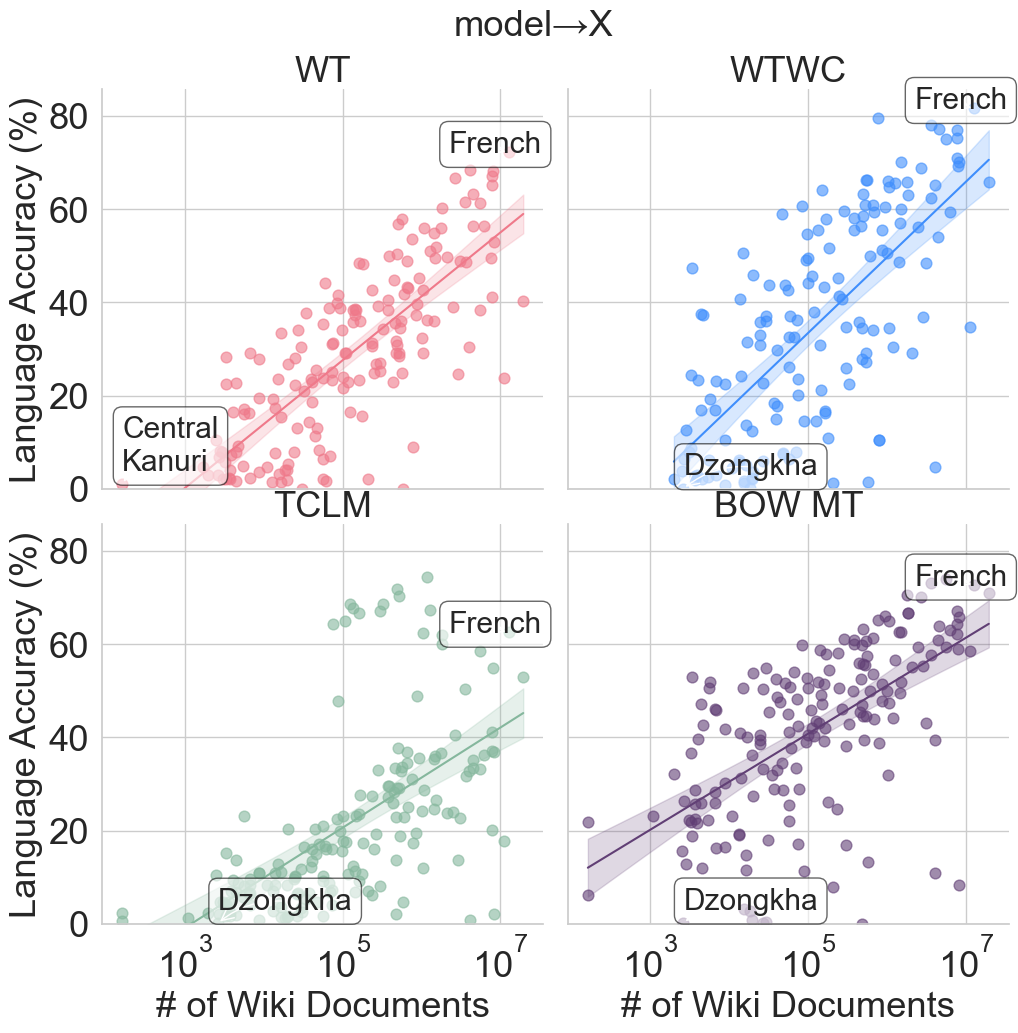}
    \end{minipage}
    \caption{Comparison of the number of Wikipedia documents\textemdash a proxy for resource level\textemdash and language performance for each task. For each language, the highest score among the six evaluated models was used. Resource levels are shown on a logarithmic scale to account for their wide range. Scatterplot labels indicate the lowest-performing low-resource language and the highest-performing high-resource language. The fitted lines in each plot depict the overall trend between resource level and performance. The shaded regions represent 95\% confidence band, which are consistently narrow and indicate the high precision of the fitted lines.}
    \label{fig:language_performance_versus_resource_level}
\end{figure*}

\autoref{fig:language_performance_versus_resource_level} compares resource level against language scores across all tasks and evaluation directions.

\subsection{Sampled Languages}
We sample 22 languages in our four tasks and display their scores in \autoref{tab:sample_langs_wt}, \autoref{tab:sample_langs_wtic}, \autoref{tab:sample_langs_fitb}, and \autoref{tab:sample_langs_bow_mt}.

\begin{table*}[!hbtp]
    \centering
    \resizebox{1\textwidth}{!}{
    \begin{tabular}{l | r r | r r | r r | r r | r r | r r}
    \toprule
    & \multicolumn{2}{c|}{\textbf{\ayaOne}} 
    & \multicolumn{2}{c|}{\textbf{\ayaTwo}} 
    & \multicolumn{2}{c|}{\textbf{\bloom}} 
    & \multicolumn{2}{c|}{\makecell{\textbf{\ttt{falcon-7b-}}\\\textbf{\ttt{instruct}}}}
    & \multicolumn{2}{c|}{\textbf{\gemma}} 
    & \multicolumn{2}{c}{\makecell{\textbf{\ttt{LLaMA-3.1-8b}}\\\textbf{\ttt{Instruct}}}} \\
    \midrule
    \textbf{Language} 
    & \textbf{\comprehension} & \textbf{\generation}
    & \textbf{\comprehension} & \textbf{\generation}
    & \textbf{\comprehension} & \textbf{\generation}
    & \textbf{\comprehension} & \textbf{\generation}
    & \textbf{\comprehension} & \textbf{\generation}
    & \textbf{\comprehension} & \textbf{\generation} \\
    \midrule
    Romanian & 55.7 & 66.6 & 63.3 & 60.9 & 35.8 & 34.4 & 32.0 & 39.8 & 38.5 & 40.5 & 49.6 & 38.6 \\
    Bulgarian & 52.3 & 48.9 & 41.3 & 42.7 & 24.9 & 5.4 & 6.9 & 8.6 & 25.0 & 10.3 & 43.0 & 4.6 \\
    Slovak & 50.0 & 27.2 & 42.2 & 50.7 & 23.4 & 15.7 & 17.1 & 11.5 & 18.4 & 18.5 & 45.2 & 58.0 \\
    Haitian & 46.6 & 39.8 & 30.2 & 27.3 & 25.2 & 14.4 & 15.4 & 12.2 & 22.8 & 5.4 & 36.0 & 25.6 \\
    Spanish & 44.7 & 55.8 & 50.4 & 61.1 & 51.2 & 68.3 & 25.4 & 62.2 & 34.6 & 47.1 & 50.0 & 62.1 \\
    Awadhi & 41.6 & 0.0 & 44.3 & 17.1 & 43.3 & 10.5 & 0.0 & 0.3 & 10.5 & 0.6 & 41.9 & 13.9 \\
    Czech & 37.8 & 28.1 & 48.0 & 49.5 & 16.0 & 10.2 & 14.2 & 11.9 & 14.2 & 23.6 & 43.8 & 41.9 \\
    Friulian & 36.7 & 27.8 & 30.8 & 27.3 & 31.8 & 20.3 & 20.5 & 27.9 & 25.5 & 14.3 & 31.2 & 15.6 \\
    Korean & 34.1 & 18.4 & 39.3 & 48.9 & 9.3 & 0.1 & 2.4 & 0.0 & 15.2 & 12.8 & 29.6 & 22.5 \\
    Sundanese & 32.7 & 34.0 & 16.4 & 23.3 & 21.1 & 20.8 & 6.5 & 2.3 & 6.5 & 7.6 & 20.7 & 16.9 \\
    Telugu & 30.8 & 34.4 & 7.0 & 4.7 & 28.9 & 30.5 & 0.0 & 0.0 & 3.0 & 0.0 & 11.4 & 26.4 \\
    Hungarian & 28.7 & 49.4 & 17.9 & 22.4 & 8.3 & 2.3 & 3.5 & 18.5 & 6.6 & 5.3 & 28.4 & 51.8 \\
    Balinese & 25.9 & 22.6 & 27.7 & 23.5 & 25.6 & 23.3 & 8.2 & 1.1 & 7.7 & 5.5 & 25.7 & 15.0 \\
    Sindhi & 23.7 & 23.8 & 20.7 & 2.4 & 11.7 & 0.0 & 2.0 & 0.0 & 7.3 & 0.6 & 33.3 & 18.8 \\
    Turkmen & 17.1 & 2.5 & 17.9 & 4.8 & 11.4 & 6.4 & 6.5 & 2.2 & 9.4 & 3.2 & 20.9 & 15.5 \\
    Pedi & 12.7 & 1.0 & 3.8 & 1.8 & 16.7 & 3.5 & 2.9 & 2.2 & 4.6 & 4.7 & 5.0 & 2.4 \\
    Sanskrit & 11.0 & 23.4 & 13.3 & 23.3 & 11.3 & 7.1 & 1.0 & 0.0 & 4.0 & 0.0 & 19.7 & 19.7 \\
    Somali & 8.2 & 22.2 & 5.4 & 12.0 & 11.7 & 2.2 & 2.2 & 0.8 & 6.4 & 1.4 & 11.6 & 6.5 \\
    Sango & 5.1 & 0.9 & 7.8 & 0.3 & 6.9 & 2.9 & 2.3 & 0.0 & 6.6 & 0.3 & 11.8 & 0.7 \\
    Nyanja & 3.4 & 9.2 & 4.0 & 0.1 & 8.1 & 5.4 & 1.0 & 0.4 & 1.7 & 0.2 & 9.7 & 4.0 \\
    Kabyle & 0.7 & 0.0 & 0.4 & 0.0 & 3.6 & 0.2 & 0.4 & 1.4 & 2.4 & 0.0 & 3.0 & 0.4 \\
    Mossi & 0.0 & 0.5 & 3.3 & 0.0 & 5.4 & 1.3 & 1.6 & 0.0 & 4.4 & 0.0 & 5.4 & 0.3 \\
    \bottomrule
    \end{tabular}
    }
    \caption{Performance on task \textbf{Word Translation} across 22 sampled languages.} %
    \label{tab:sample_langs_wt}
\end{table*}

\begin{table*}[!hbtp]
    \centering
    \resizebox{1\textwidth}{!}{
    \begin{tabular}{l | r r | r r | r r | r r | r r | r r}
        \toprule
        & \multicolumn{2}{c|}{\textbf{\ayaOne}} & \multicolumn{2}{c|}{\textbf{\ayaTwo}} & \multicolumn{2}{c|}{\textbf{\bloom}} & \multicolumn{2}{c|}{\makecell{\textbf{\ttt{falcon-7b-}}\\\textbf{\ttt{instruct}}}} & \multicolumn{2}{c}{\textbf{\gemma}} & \multicolumn{2}{|c}{\makecell{\textbf{\ttt{LLaMA-3.1-8b}}\\\textbf{\ttt{Instruct}}}} \\
        \midrule
        \textbf{Language} & \multicolumn{1}{c}{\textbf{\comprehension}} & \multicolumn{1}{c|}{\textbf{\generation}} & \multicolumn{1}{c}{\textbf{\comprehension}} & \multicolumn{1}{c|}{\textbf{\generation}} & \multicolumn{1}{c}{\textbf{\comprehension}} & \multicolumn{1}{c|}{\textbf{\generation}} & \multicolumn{1}{c}{\textbf{\comprehension}} & \multicolumn{1}{c|}{\textbf{\generation}} & \multicolumn{1}{c}{\textbf{\comprehension}} & \multicolumn{1}{c|}{\textbf{\generation}} & {\textbf{\comprehension}} & \multicolumn{1}{c}{\textbf{\generation}} \\ 
        \midrule
        Spanish & 78.8 & 69.3 & 81.1 & 64.0 & 30.3 & 65.0 & 83.4 & 65.0 & 77.0 & 34.9 & 51.7 & 59.7 \\
        Bulgarian & 76.1 & 60.8 & 70.5 & 33.1 & 7.8 & 3.2 & 29.4 & 9.8 & 46.5 & 1.6 & 41.3 & 5.8 \\
        Slovak & 70.0 & 66.2 & 78.0 & 42.5 & 6.5 & 16.9 & 38.5 & 17.3 & 63.0 & 15.3 & 55.2 & 21.0 \\
        Korean & 68.6 & 31.2 & 63.4 & 48.5 & 22.0 & 1.3 & 11.8 & 0.1 & 48.4 & 3.9 & 54.7 & 0.6 \\
        Czech & 68.2 & 70.1 & 69.8 & 62.6 & 5.3 & 7.2 & 35.9 & 27.9 & 48.2 & 16.8 & 60.1 & 31.1 \\
        Hungarian & 64.1 & 60.1 & 59.4 & 24.1 & 1.3 & 11.1 & 8.7 & 6.7 & 36.6 & 5.1 & 46.5 & 23.9 \\
        Romanian & 63.6 & 68.9 & 65.4 & 59.3 & 16.9 & 8.1 & 58.8 & 29.8 & 54.0 & 13.1 & 32.3 & 33.1 \\
        Haitian & 61.2 & 60.6 & 48.4 & 2.6 & 7.8 & 7.9 & 27.9 & 10.0 & 27.1 & 2.0 & 24.7 & 14.3 \\
        Sundanese & 58.2 & 54.8 & 32.1 & 12.6 & 3.4 & 4.3 & 13.3 & 2.6 & 16.8 & 0.8 & 19.3 & 5.6 \\
        Turkmen & 45.2 & 13.6 & 33.8 & 4.7 & 1.5 & 1.1 & 6.0 & 4.8 & 16.3 & 0.7 & 27.8 & 8.3 \\
        Pedi & 44.9 & 0.6 & 6.5 & 0.3 & 9.0 & 3.7 & 6.5 & 2.8 & 11.7 & 0.5 & 5.9 & 4.5 \\
        Balinese & 43.9 & 14.9 & 25.3 & 8.6 & 1.2 & 2.0 & 12.8 & 2.4 & 16.7 & 1.3 & 21.9 & 4.5 \\
        Friulian & 42.6 & 22.4 & 43.8 & 12.4 & 3.2 & 7.9 & 28.6 & 18.7 & 21.9 & 6.5 & 22.7 & 16.8 \\
        Awadhi & 39.2 & 0.0 & 39.1 & 6.2 & 2.4 & 0.0 & 2.6 & 0.0 & 27.8 & 0.0 & 32.2 & 0.0 \\
        Sindhi & 32.2 & 32.5 & 30.8 & 1.4 & 1.1 & 0.5 & 3.8 & 0.5 & 16.6 & 0.0 & 40.4 & 0.6 \\
        Sango & 29.8 & 1.6 & 6.6 & 1.0 & 1.9 & 0.7 & 6.7 & 0.9 & 11.9 & 0.0 & 3.1 & 2.1 \\
        Sanskrit & 29.3 & 1.6 & 27.2 & 23.4 & 0.8 & 0.0 & 6.1 & 5.9 & 22.2 & 0.2 & 23.2 & 1.9 \\
        Nyanja & 28.3 & 37.3 & 7.4 & 2.4 & 3.0 & 18.8 & 4.2 & 3.2 & 5.3 & 0.9 & 10.0 & 6.3 \\
        Telugu & 21.0 & 22.4 & 10.9 & 2.1 & 2.4 & 10.3 & 1.7 & 0.1 & 14.1 & 0.0 & 6.7 & 6.9 \\
        Somali & 9.7 & 33.1 & 15.9 & 3.1 & 2.2 & 0.7 & 3.9 & 1.6 & 12.8 & 0.1 & 6.9 & 2.9 \\
        Mossi & 3.7 & 0.4 & 3.4 & 0.5 & 1.0 & 0.1 & 3.1 & 0.3 & 8.0 & 0.1 & 0.8 & 4.5 \\
        Kabyle & 2.0 & 3.4 & 4.1 & 0.6 & 1.6 & 0.4 & 1.6 & 3.6 & 9.2 & 0.1 & 2.6 & 5.2 \\
        \bottomrule
    \end{tabular}}
    \caption{Performance on task \textbf{Word Translation-in-Context} across 22 sampled languages.} %
    \label{tab:sample_langs_wtic}
\end{table*}

\begin{table*}[!hbtp]
    \centering
    \resizebox{1\textwidth}{!}{
    \begin{tabular}{l | r r | r r | r r | r r | r r | r r }
        \toprule
        & \multicolumn{2}{c|}{\textbf{\ayaOne}} & \multicolumn{2}{c|}{\textbf{\ayaTwo}} & \multicolumn{2}{c|}{\textbf{\bloom}} & \multicolumn{2}{c|}{\makecell{\textbf{\ttt{falcon-7b-}}\\\textbf{\ttt{instruct}}}} & \multicolumn{2}{c|}{\textbf{\gemma}} & \multicolumn{2}{c}{\makecell{\textbf{\ttt{LLaMA-3.1-8b}}\\\textbf{\ttt{Instruct}}}}\\
        \midrule
        \textbf{Language} & \multicolumn{1}{c}{\textbf{\comprehension}} & \multicolumn{1}{c|}{\textbf{\generation}} & \multicolumn{1}{c}{\textbf{\comprehension}} & \multicolumn{1}{c|}{\textbf{\generation}} & \multicolumn{1}{c}{\textbf{\comprehension}} & \multicolumn{1}{c|}{\textbf{\generation}} & \multicolumn{1}{c}{\textbf{\comprehension}} & \multicolumn{1}{c|}{\textbf{\generation}} & \multicolumn{1}{c}{\textbf{\comprehension}} & \multicolumn{1}{c|}{\textbf{\generation}} & \multicolumn{1}{c}{\textbf{\comprehension}} & \multicolumn{1}{c}{\textbf{\generation}} \\ 
        \midrule
        Romanian & 47.3 & 38.4 & 49.1 & 40.6 & 48.5 & 4.3 & 32.8 & 7.5 & 39.7 & 18.9 & 50.7 & 38.5 \\
        Telugu & 46.7 & 24.3 & 17.9 & 1.8 & 64.3 & 68.5 & 14.3 & 0.3 & 18.1 & 2.2 & 45.4 & 19.7 \\
        Bulgarian & 46.6 & 36.8 & 41.1 & 8.2 & 44.7 & 2.5 & 15.1 & 0.8 & 39.7 & 12.8 & 49.9 & 29.1 \\
        Sindhi & 46.5 & 21.0 & 18.4 & 1.5 & 37.5 & 1.1 & 12.5 & 0.2 & 16.0 & 1.2 & 40.3 & 10.1 \\
        Czech & 46.3 & 33.2 & 47.6 & 35.9 & 38.9 & 2.4 & 23.6 & 2.6 & 39.9 & 16.1 & 49.2 & 33.5 \\
        Awadhi & 45.8 & 16.6 & 39.7 & 15.8 & 63.5 & 23.0 & 13.8 & 0.2 & 27.0 & 6.1 & 44.5 & 19.2 \\
        Sundanese & 45.7 & 15.6 & 27.6 & 5.3 & 46.5 & 5.0 & 18.6 & 1.1 & 21.9 & 3.8 & 41.2 & 8.6 \\
        Slovak & 45.4 & 33.7 & 44.1 & 14.4 & 37.6 & 2.1 & 19.6 & 1.6 & 35.1 & 10.9 & 47.1 & 26.0 \\
        Haitian & 44.6 & 22.4 & 21.9 & 3.2 & 39.9 & 2.3 & 16.8 & 1.2 & 18.6 & 2.4 & 37.3 & 8.1 \\
        Spanish & 43.4 & 36.4 & 44.4 & 37.3 & 55.7 & 54.8 & 40.7 & 29.5 & 41.5 & 31.8 & 46.0 & 38.4 \\
        Hungarian & 43.3 & 26.5 & 31.1 & 4.2 & 31.7 & 0.9 & 14.2 & 0.5 & 28.0 & 5.5 & 46.8 & 26.3 \\
        Korean & 41.3 & 20.1 & 42.4 & 22.7 & 40.7 & 1.2 & 14.8 & 0.3 & 35.7 & 11.2 & 44.6 & 18.5 \\
        Friulian & 40.9 & 4.4 & 33.6 & 4.8 & 46.2 & 3.0 & 22.1 & 1.5 & 25.5 & 3.1 & 41.4 & 11.0 \\
        Balinese & 39.8 & 12.3 & 28.4 & 8.2 & 44.9 & 8.2 & 19.0 & 1.8 & 24.4 & 6.4 & 38.0 & 8.9 \\
        Pedi & 38.3 & 6.8 & 14.3 & 1.7 & 57.9 & 8.4 & 13.9 & 1.1 & 14.4 & 2.5 & 20.3 & 3.3 \\
        Somali & 36.7 & 10.1 & 15.8 & 2.0 & 30.6 & 1.0 & 14.5 & 0.7 & 14.5 & 1.4 & 22.1 & 2.3 \\
        Nyanja & 35.6 & 8.7 & 15.0 & 1.6 & 52.8 & 7.0 & 14.6 & 1.1 & 14.9 & 1.7 & 22.1 & 2.9 \\
        Turkmen & 35.4 & 2.4 & 22.2 & 1.5 & 31.1 & 0.5 & 15.4 & 0.2 & 17.1 & 0.9 & 31.2 & 3.5 \\
        Sanskrit & 30.6 & 2.5 & 19.9 & 1.7 & 41.2 & 2.4 & 13.5 & 0.1 & 18.6 & 1.0 & 32.0 & 3.9 \\
        Mossi & 12.5 & 1.1 & 14.2 & 0.9 & 34.1 & 1.2 & 14.2 & 0.7 & 14.4 & 1.1 & 16.9 & 0.9 \\
        Sango & 11.9 & 2.1 & 11.9 & 1.5 & 31.9 & 2.4 & 11.5 & 0.9 & 11.0 & 1.8 & 13.5 & 1.8 \\
        Kabyle & 10.6 & 1.5 & 11.4 & 0.7 & 26.6 & 0.8 & 11.2 & 0.4 & 11.5 & 1.0 & 14.6 & 1.4 \\
        \bottomrule
    \end{tabular}}
    \caption{Performance on task \textbf{Translation-Conditioned Language Modeling} across 22 sampled languages.} %
    \label{tab:sample_langs_fitb}
\end{table*}

\begin{table*}[!hbtp]
    \centering
    \resizebox{1\textwidth}{!}{
    \begin{tabular}{l | r r | r r | r r | r r | r r | r r }
        \toprule
        & \multicolumn{2}{c|}{\textbf{\ayaOne}} & \multicolumn{2}{c|}{\textbf{\ayaTwo}} & \multicolumn{2}{c|}{\textbf{\bloom}} & \multicolumn{2}{c|}{\makecell{\textbf{\ttt{falcon-7b-}}\\\textbf{\ttt{instruct}}}} & \multicolumn{2}{c|}{\textbf{\gemma}} & \multicolumn{2}{c}{\makecell{\textbf{\ttt{LLaMA-3.1-8b}}\\\textbf{\ttt{Instruct}}}} \\
        \midrule
        \textbf{Language} & \multicolumn{1}{c}{\textbf{\comprehension}} & \multicolumn{1}{c|}{\textbf{\generation}} & \multicolumn{1}{c}{\textbf{\comprehension}} & \multicolumn{1}{c|}{\textbf{\generation}} & \multicolumn{1}{c}{\textbf{\comprehension}} & \multicolumn{1}{c|}{\textbf{\generation}} & \multicolumn{1}{c}{\textbf{\comprehension}} & \multicolumn{1}{c|}{\textbf{\generation}} & \multicolumn{1}{c}{\textbf{\comprehension}} & \multicolumn{1}{c|}{\textbf{\generation}} & \multicolumn{1}{c}{\textbf{\comprehension}} & \multicolumn{1}{c}{\textbf{\generation}} \\ 
        \midrule
        Spanish & 87.6 & 57.0 & 88.3 & 58.9 & 9.4 & 32.2 & 79.1 & 30.9 & 85.5 & 51.6 & 79.7 & 51.9 \\
        Slovak & 84.5 & 57.5 & 82.8 & 45.2 & 0.8 & 5.5 & 43.1 & 13.4 & 70.9 & 30.9 & 77.6 & 44.0 \\
        Romanian & 82.8 & 63.5 & 85.6 & 70.2 & 4.8 & 5.4 & 57.0 & 20.6 & 74.4 & 45.1 & 78.7 & 56.2 \\
        Czech & 81.6 & 57.6 & 85.9 & 62.6 & 0.7 & 4.7 & 47.9 & 13.5 & 71.8 & 37.1 & 76.0 & 46.7 \\
        Bulgarian & 78.9 & 61.4 & 79.0 & 27.7 & 1.1 & 0.7 & 9.1 & 2.1 & 72.7 & 16.9 & 73.4 & 37.5 \\
        Korean & 76.7 & 35.1 & 80.5 & 43.1 & 5.9 & 1.0 & 21.4 & 2.9 & 62.9 & 18.6 & 71.4 & 27.1 \\
        Friulian & 76.0 & 31.7 & 67.3 & 40.9 & 0.9 & 2.5 & 44.6 & 23.0 & 56.9 & 25.9 & 67.5 & 41.9 \\
        Haitian & 75.7 & 59.7 & 49.9 & 23.0 & 0.2 & 1.6 & 34.0 & 22.4 & 44.7 & 25.6 & 65.5 & 35.7 \\
        Telugu & 74.6 & 42.8 & 45.3 & 3.8 & 3.5 & 9.1 & 2.7 & 1.7 & 44.4 & 3.7 & 23.0 & 21.2 \\
        Hungarian & 73.9 & 51.9 & 61.3 & 22.3 & 0.3 & 2.9 & 29.2 & 11.0 & 57.1 & 23.7 & 64.7 & 47.8 \\
        Sundanese & 73.7 & 52.8 & 57.6 & 50.8 & 0.0 & 5.5 & 34.9 & 16.9 & 49.0 & 24.4 & 64.5 & 35.4 \\
        Balinese & 71.2 & 45.0 & 56.1 & 52.5 & 0.4 & 3.6 & 35.8 & 18.0 & 50.0 & 35.9 & 62.2 & 43.5 \\
        Sanskrit & 66.2 & 15.6 & 51.3 & 17.2 & 0.1 & 0.4 & 3.3 & 1.4 & 47.6 & 7.5 & 55.9 & 11.4 \\
        Turkmen & 65.9 & 11.1 & 44.0 & 12.9 & 0.0 & 0.5 & 23.8 & 6.9 & 36.0 & 8.5 & 56.2 & 14.7 \\
        Awadhi & 65.9 & 3.6 & 76.2 & 50.6 & 0.5 & 0.6 & 4.0 & 1.9 & 56.8 & 19.3 & 71.2 & 32.9 \\
        Nyanja & 64.0 & 42.7 & 32.2 & 14.9 & 0.0 & 0.6 & 26.1 & 14.3 & 34.2 & 14.6 & 48.3 & 19.0 \\
        Pedi & 61.8 & 21.3 & 38.9 & 18.7 & 0.8 & 0.6 & 32.7 & 16.0 & 37.7 & 19.1 & 45.9 & 23.1 \\
        Sindhi & 58.3 & 41.2 & 33.7 & 3.7 & 0.0 & 0.4 & 2.3 & 2.3 & 31.9 & 3.9 & 60.3 & 19.0 \\
        Somali & 53.7 & 39.5 & 34.1 & 17.5 & 0.1 & 0.8 & 21.3 & 10.3 & 28.7 & 11.0 & 46.3 & 17.8 \\
        Sango & 50.1 & 32.2 & 34.5 & 22.4 & 0.1 & 6.7 & 29.2 & 20.0 & 39.4 & 20.7 & 43.4 & 29.0 \\
        Mossi & 34.2 & 14.4 & 28.5 & 17.0 & 0.0 & 0.6 & 25.2 & 10.0 & 30.9 & 14.1 & 34.2 & 16.0 \\
        Kabyle & 19.2 & 9.9 & 19.5 & 8.8 & 0.0 & 0.5 & 16.0 & 7.7 & 20.8 & 9.0 & 31.3 & 11.5 \\
        \bottomrule
    \end{tabular}}
    \caption{Performance on task \textbf{Bag-of-Words Machine Translation} across 22 sampled languages.} %
    \label{tab:sample_langs_bow_mt}
\end{table*}

\newpage 
\subsection{Language Score Distribution}
\begin{figure*}
    \centering
    \includegraphics[width=1\linewidth]{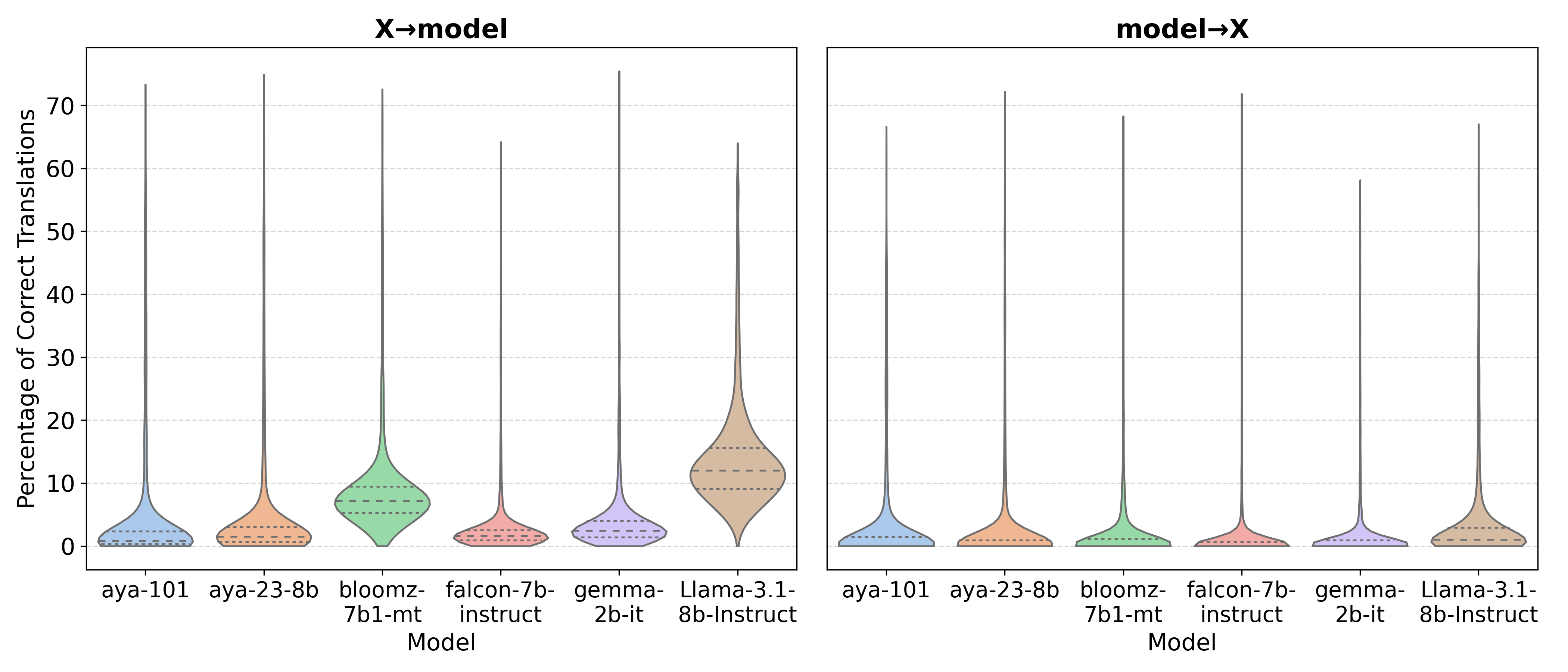}
    \caption{Model-wise performance distribution for the task \textbf{Word Translation}. Each violin depicts the distribution of scores across evaluated languages. Dotted lines indicate the first, second, and third quartiles of this distribution.}
    \label{fig:wt_general_correct}
\end{figure*}

\begin{figure*}
    \centering
    \includegraphics[width=1\linewidth]{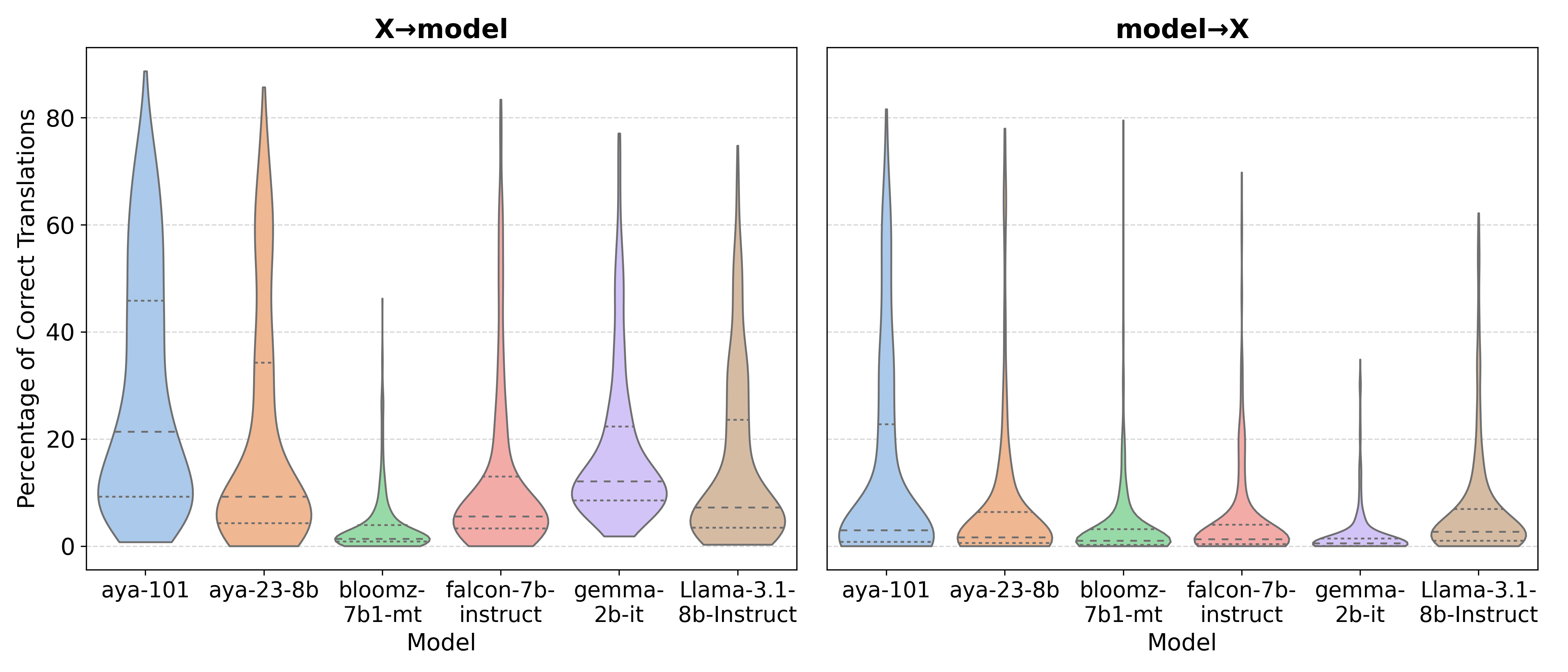}
    \caption{Model-wise performance distribution for the task \textbf{Word Translation with Context}. Each violin depicts the distribution of scores across evaluated languages. Dotted lines indicate the first, second, and third quartile of this distribution.}
    \label{fig:wtic_general_correct}
\end{figure*}

\begin{figure*}
    \centering
    \includegraphics[width=1\linewidth]{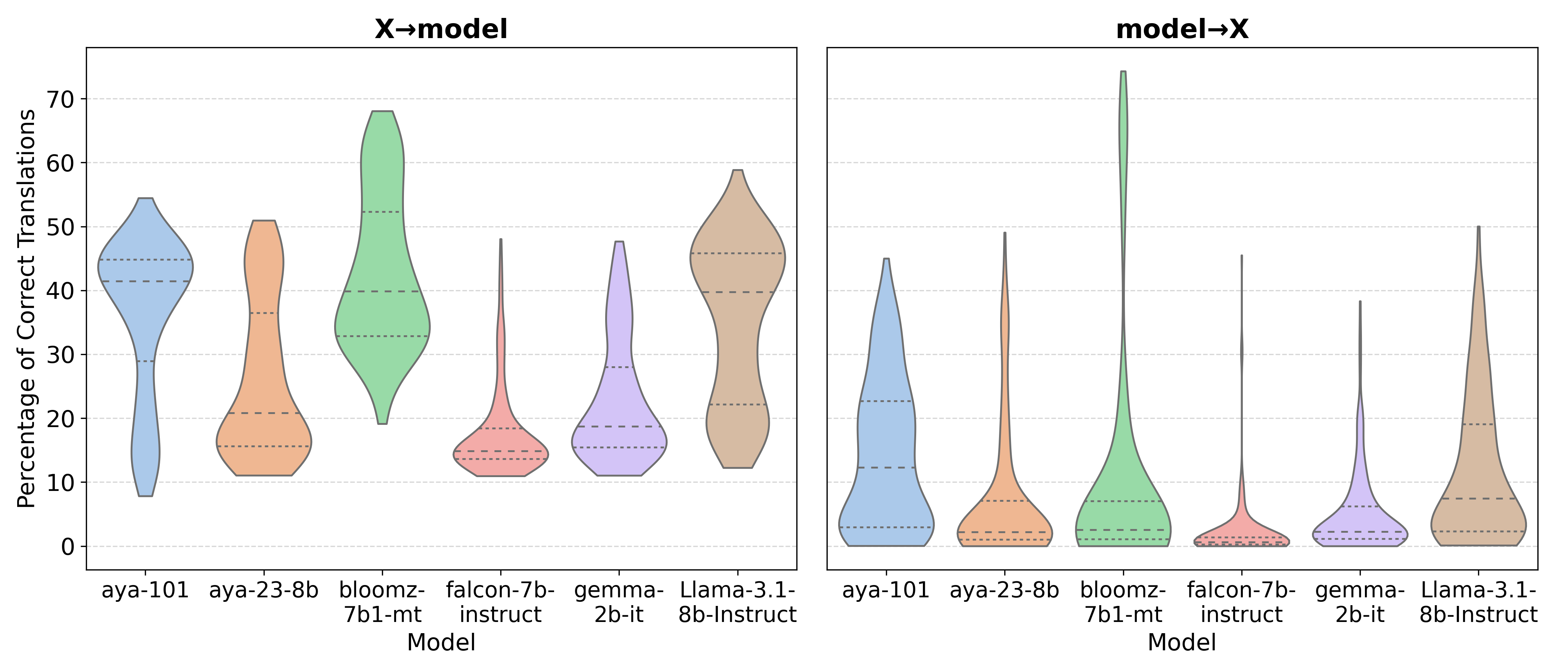}
    \caption{Model-wise performance distribution for the task \textbf{Translation-Conditioned Language Modeling}. Each violin depicts the distribution of scores across evaluated languages. Dotted lines indicate the first, second, and third quartile of this distribution.}
    \label{fig:fitb_general_correct}
\end{figure*}

\begin{figure*}
    \centering
    \includegraphics[width=1\linewidth]{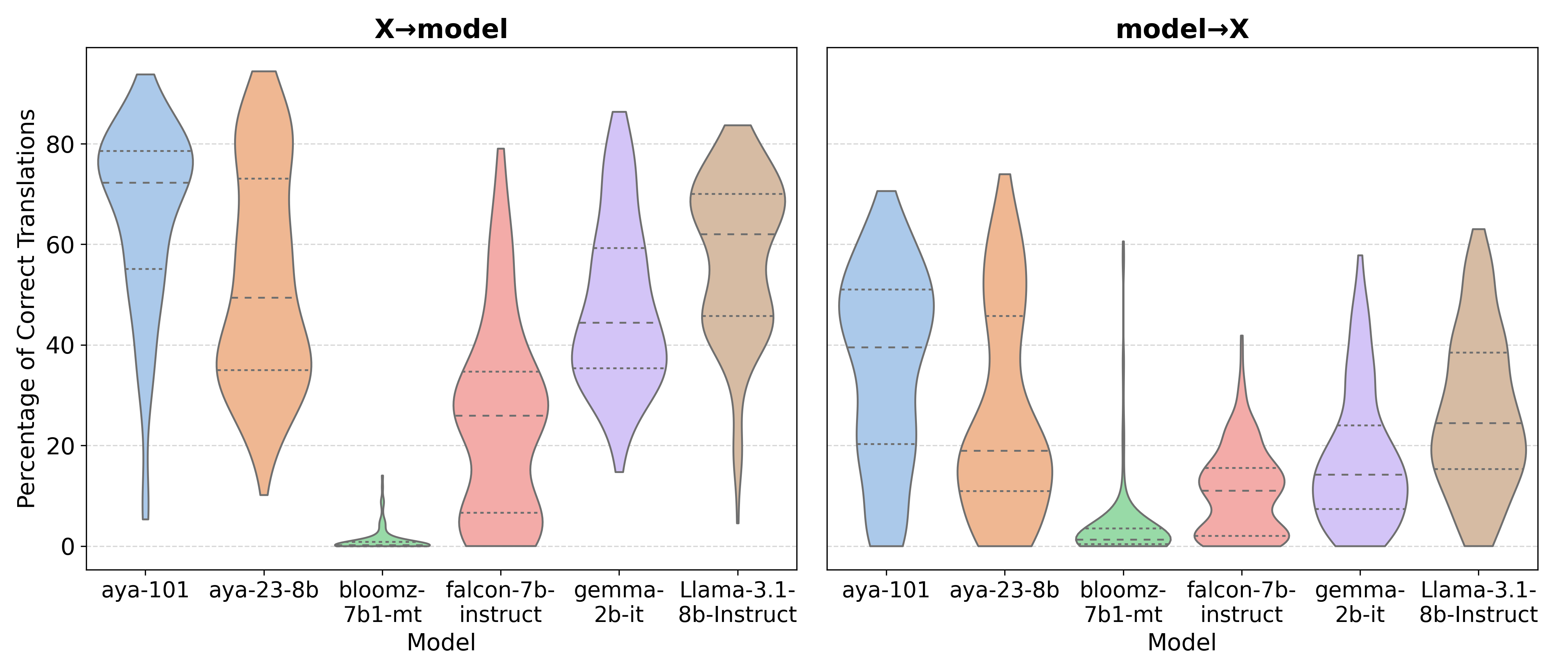}
    \caption{Model-wise performance distribution for the task \textbf{Bag-of-Words Machine Translation}. Each violin depicts the distribution of scores across evaluated languages. Dotted lines indicate the first, second, and third quartile of this distribution.}
    \label{fig:bow_mt_general_correct}
\end{figure*}

\autoref{fig:wt_general_correct}, \autoref{fig:wtic_general_correct}, \autoref{fig:fitb_general_correct}, and \autoref{fig:bow_mt_general_correct} outline the distribution of language scores for each task.

You may notice that \bloom~ performs especially badly in \autoref{fig:bow_mt_general_correct}. Interestingly, the model average is higher in \generation~ than in \comprehension. The model performs poorly even with HRLs, receiving a score of 9.4\% for the Spanish$\rightarrow$English translations (see \autoref{tab:sample_langs_bow_mt}). The model \bloom~ had difficulty following instructions, often echoing the prompt. For example, \bloom~ echoes the source sentence when tasked with translating a Swedish sentence:

\begin{quote}
    \textbf{Prompt} \\
    Translate into English: ``Vi har nu 4 månader gamla möss som har blivit kvitt sin diabetes'', tillade han.\\
    \\
    \textbf{Model Response:} \\
    Translate into English: ``Vi har nu 4 månader gamla möss som har blivit kvitt sin diabetes'', tillade han.
\end{quote}

\newpage
\section{Sampling} \label{sec:sampling}
Due to the large size of our dataset and limited compute, we evaluated only a sample of existing data. We explain what this means in each task.

\begin{itemize}
    \item Word Translation: We randomly sample 300 entries from the translation lexicon should more than 300 entries exist.
    \item Word Translation With Context: We prompt a model until we have evaluated 300 unique words.
    \item Translation-Conditioned Language Modeling: We prompt the model on words from the first 300 sentences in FLORES+.
    \item Bag-of-Words Machine Translation: Similarly to \FillInTheBlankAbbreviation, we prompt the model on words from the first 300 sentences in FLORES+.

\end{itemize}

\section{Evaluation Details} \label{sec:evaluation_details}
We used A40s A100s, and A6000s to run evaluation on \WordTranslationAbbreviation, \WordTranslationInContextAbbreviation, \FillInTheBlankAbbreviation, and \BagOfWordsMachineTranslationAbbreviation. We discuss the GPU compute hours in more detail.

\subsection{Compute}
\paragraph{Word Translation} We conducted evaluation for 2,746 languages $\times$ 2 evaluation directions $\times$ 6 model $=$ 32,952 evaluations. Each run takes approximately 6 minutes, resulting in 3,295.2 GPU hours.

\paragraph{Word Translation with Context} We conducted evaluation for 525 languages $\times$ 2 evaluation directions $\times$ 6 models $=$ 6,300 evaluations. Each run takes approximately 40 minutes, resulting in 4,200 GPU hours.

\paragraph{Translation-Conditioned Language Modeling} We conducted evaluation for 211 languages $\times$ 2 evaluation directions $\times$ 6 models $=$ 2,532 evaluations. Each run takes approximately 6 minutes, resulting in 253.2 GPU hours.

\paragraph{Bag-of-Words Machine Translation} We conducted evaluation for 211 languages $\times$ 2 evaluation directions $\times$ 6 models $=$ 2,532 evaluations. Each run takes approximately 3 minutes, resulting in 126.6 GPU hours.

\subsection{Evaluation}
We tested our models on \ttt{devtest} splits of the FLORES+ dataset and version \ttt{v3.1} from GLOTLID. We also used \textsc{BLEU} (HuggingFace \ttt{evaluate} wrapper), and \textsc{WordNet} from \ttt{nltk.corpus}.

\begin{table*}[!hbtp]
    \centering
    \small
        \begin{tabular}{l | r | c | c | l | l }
            \toprule
            \textbf{Model} &
                \makecell{\textbf{Languages}\\\textbf{Supported}} &
                \makecell{\textbf{Release}\\\textbf{Year}} & \textbf{Architecture} & \makecell[l]{\textbf{Training Data}\\\textbf{Mixture}} &  \textbf{Rationale} \\
            \midrule
            \ayaOne~ (13B) & 101 & 2024 & \makecell{encoder-\\decoder} & \makecell[l]{multilingual templates,\\human annotations,\\synthetic data,\\machine translation} & \makecell[l]{trained on many\\low-resource\\languages} \\
            \midrule
            \ayaTwo & 23 & 2024 & decoder & \makecell[l]{multilingual templates,\\human annotations,\\synthetic data,\\machine translation} & \makecell[l]{outperforms \ayaOne\\across 23 covered\\languages} \\
            \midrule
            \bloom & 45 & 2023 & decoder & \makecell[l]{machine translation,\\simplification,\\program synthesis,\\code datasets} & \makecell[l]{reputed for strong\\cross-lingual\\generalization} \\
            \midrule
            \falcon & 11 & 2023 & decoder & \makecell[l]{instruct and chat\\datasets} & \makecell[l]{multilingual in\\high-resource\\languages (\ttt{fra})} \\ %
            \midrule
            \gemma & 1 & 2024 & decoder & \makecell[l]{web documents, code,\\math} & \makecell[l]{English-trained\\model (study control)} \\
            \midrule
            \llama & 8 & 2024 & decoder & public online data & \makecell[l]{multilingual in mid-\\and high-resource\\languages} \\ %
            \bottomrule
        \end{tabular}
    \caption{Overview of the language models evaluated in this study, the number of languages each model supports, model release year, basic model architecture, datasets used to train and finetune the model, as well as the rationale for why the model was selected.}
    \label{tab:model_overview}
\end{table*}

\end{document}